\documentclass[lettersize,journal]{IEEEtran}
\usepackage{amsmath,amsfonts}
\usepackage{algorithmic}
\usepackage{algorithm}
\usepackage{array}
\usepackage[caption=false,font=normalsize,labelfont=sf,textfont=sf]{subfig}
\usepackage{textcomp}
\usepackage{stfloats}
\usepackage{url}
\usepackage{verbatim}
\usepackage{graphicx}
\usepackage{cite}
\usepackage{times}
\usepackage{soul}
\usepackage{url}
\usepackage[hidelinks]{hyperref}
\usepackage{graphicx}
\usepackage{amsthm}
\usepackage[caption=false,font=normalsize,labelfont=sf,textfont=sf]{subfig}

\usepackage{multirow}
\usepackage{amssymb}
\usepackage{color}
\usepackage{xcolor}
\usepackage{colortbl}

\newcommand{\red}[1]{{\color{red}#1}}
\newcommand{\blue}[1]{{\color{blue}#1}}

\newcommand{\gray}[1]{{\color{gray}#1}}

\usepackage{pict2e,picture} 
\usepackage{tikz}   
\usepackage{ulem}
\usepackage{pifont} 

\newcommand*\numcircledmod[1]{\raisebox{.5pt}{\textcircled{\raisebox{-.9pt} {#1}}}}

\newsavebox\CBox
\newlength\CLength
\def\numcircledpict#1{\sbox\CBox{#1}%
  \ifdim\wd\CBox>\ht\CBox \CLength=\wd\CBox\else\CLength=\ht\CBox\fi
    \makebox[1.5\CLength]{\makebox(0,1.5\CLength){\put(0,0){\circle{1.5\CLength}}}%
    \makebox(0,1.5\CLength){\put(-.5\wd\CBox,0){#1}}}}

\newtheorem{definition}{Definition}

\begin{document}

\title{A Hybrid Self-Supervised Learning Framework for Vertical Federated Learning}

\author{
Yuanqin He,
Yan Kang\footnote{Contact Author},
Xinyuan Zhao,
Jiahuan Luo,
Lixin Fan,
Yuxing Han,
Qiang Yang$~\IEEEmembership{Fellow,~IEEE}$
}%
\markboth{Journal of \LaTeX\ Class Files,~Vol.~14, No.~8, June~2023}%
{Shell \MakeLowercase{\textit{et al.}}: A Sample Article Using IEEEtran.cls for IEEE Journals}


\maketitle

\begin{abstract}
Vertical federated learning (VFL), a variant of Federated Learning (FL), has recently drawn increasing attention as the VFL matches the enterprises' demands of leveraging more valuable features to achieve better model performance. However, conventional VFL methods may run into data deficiency as they exploit only aligned and labeled samples (belonging to different parties), leaving often the majority of unaligned and unlabeled samples unused. The data deficiency hampers the effort of the federation. 

In this work, we propose a Federated Hybrid Self-Supervised Learning framework, named FedHSSL, that utilizes cross-party views (i.e., dispersed features) of samples aligned among parties and local views (i.e., augmentation) of unaligned samples within each party to improve the representation learning capability of the VFL joint model. FedHSSL further exploits invariant features across parties to boost the performance of the joint model through partial model aggregation. FedHSSL, as a framework, can work with various representative SSL methods. We empirically demonstrate that FedHSSL methods outperform baselines by large margins. We provide an in-depth analysis of FedHSSL regarding label leakage, which is rarely investigated in existing self-supervised VFL works. The experimental results show that, with proper protection, FedHSSL achieves the best privacy-utility trade-off against the state-of-the-art label inference attack compared with baselines. Code is available at \url{https://github.com/jorghyq2016/FedHSSL}.
\end{abstract}

\begin{IEEEkeywords}
Vertical federated learning, self-supervised learning, privacy preservation, neural network.
\end{IEEEkeywords}


\section{Introduction}
\label{sec:introduction}
Federated learning (FL) enables independent parties to build machine learning models collaboratively without sharing private data~\cite{mcmahan2017communicationefficient,yang2019federated}. This makes FL a practical solution to tackle data silo issues while complying with increasingly strict legal and regulatory constraints enforced on user privacy, such as the General Data Protection Regulation (GDPR). ~\cite{yang2019federated} categorizes FL into  \textit{Horizontal} FL (HFL) and \textit{Vertical} FL (VFL). HFL typically involves a large number of parties that have different samples but share the same feature space, while VFL involves several parties that own distinct features of the same set of samples. Recently, VFL has drawn increasing attention as the VFL matches the enterprises’ demands of leveraging more valuable features to achieve better model performance without jeopardizing data privacy. e.g., VFL has been widely deployed in industries such as finance~\cite{kang2021privacy} and advertisement~\cite{tan2020rec}.

However, VFL has two critical limitations. One is the deficiency of labeled samples. For example, positive labels are costly in the credit risk assessment because they are available only when customers either complete their repayment or default, which may take a few years. Another limitation is the deficiency of aligned samples. When participating parties have quite different customer bases, their aligned samples are likely to be very limited. To address these two limitations,~\cite{kangyan2022fedcvt} proposed a federated cross-view approach that leverages the aligned samples to estimate missing features and labels, which in turn is utilized for training the joint VFL model. This approach essentially relies on aligned samples and is conducted in a supervised learning manner. Recently, self-supervised learning (SSL) has been introduced to HFL, aiming to improve the representation learning capability of the global model on label deficiency scenarios~\cite{zhuang2022divergenceaware,chu2021privacypreserving}, while the research on integrating SSL into VFL is understudied. Existing SSL works in VFL either solely used local unlabeled data~ \cite{castiglia2022lssl, feng2022vertical} without considering cross-party views of the aligned samples or only focused on aligned unlabeled sample~\cite{li2022achieving}, but failed to exploit each party's local data. Besides, although SSL does not involve labels, sample/feature alignment may result in the leakage of label information. Existing SSL-based VFL works rarely studied the impact of SSL on label leakage.
 
To fill these gaps, we propose FedHSSL, a \underline{Fed}erated \underline{H}ybrid \underline{S}elf-\underline{S}upervised \underline{L}earning framework (illustrated in Fig.~\ref{fig:overview}). FedHSSL simultaneously exploits (i) cross-party views (i.e., dispersed features) of samples aligned among parties and (ii) local views (i.e., augmentations) of samples within each party, and aggregates (iii) invariant features shared among parties, aiming to improve the overall performance of the final joint model. Furthermore, we analyze the label leakage of both the pretraining and fine-tuning phases of FedHSSL and investigate the protection against the label inference attack on FedHSSL. Our contributions are as follows: 
\begin{itemize}
    \item We propose a federated hybrid SSL framework that takes advantage of all available data through SSL and partial model aggregation to address the data deficiency issue in VFL. Experimental results show that FedHSSL methods outperform baselines by large margins on four datasets. The ablation study demonstrates the effectiveness of each step involved in FedHSSL in improving the performance of the VFL joint model.
    \item We analyze the label leakage issue of FedHSSL. This is one of the first attempts to study label leakage of pretrained models in VFL. Experimental results demonstrate that FedHSSL achieves a better privacy-utility trade-off than baselines. 

\end{itemize}

\section{Related Works}
\label{sec:relatedworks} 

\subsection{Vertical Federated Learning (VFL)} 
VFL aims to build a joint machine learning model using features dispersed among parties while protecting privacy~\cite{yang2019federatedb}. In recent years, the literature has presented various algorithms in the VFL setting. ~\cite{hardy2017private} proposed vertical logistic regression (VLR) using homomorphic encryption (HE) to protect data privacy. ~\cite{chen2021sshe} further enhanced the privacy-preserving capability of VLR by employing a hybrid strategy combining HE and secret sharing (SS). ~\cite{cheng2021secureboost} proposed the SecureBoost, a VFL version of XGBoost, that leverages HE to protect the parameters exchanged among parties. To tackle the data deficiency issue of VFL,~\cite{liu2020secure} integrated transfer learning into VFL to help the target party predict labels. ~\cite{kangyan2022fedcvt} applied a semi-supervised learning method to estimate missing features and labels for further training.

\subsection{Self (Semi)-Supervised Learning in VFL} \label{sec:rel_ssl}


With the success of contrastive learning in computer vision, it gradually dominates self-supervised learning (SSL)~\cite{bachman2019learninga, chen2020simple}. While several works applied SSL to HFL to address non-IID~\cite{li2021modelcontrastive,mu2021fedproc} or label deficiency issues ~\cite{zhang2020federated,zhuang2021collaborative,zhuang2022divergenceaware,he2021ssfl}, the research on integrating SSL into VFL is limited.~\cite{castiglia2022lssl, feng2022vertical} pretrained participating parties' local models leveraging their unaligned local samples without considering aligned samples. \cite{li2022achieving} used aligned samples for learning discriminative representations but did not use unlabeled local samples.~\cite{kangyan2022fedcvt, yang2022multiview} exploited semi-supervised learning techniques to predict pseudo labels of unaligned samples and estimate missing features to boost the performance of VFL joint models. Table \ref{table:SSL_works} briefly summarizes these works. 

Several VFL works aim to build a local predictor for one party instead of a VFL joint model. For example, the goal of ~\cite{huang2023vertical,ren2022improvingb,li2022vertical} is to train a local predictor for the active party for addressing the efficiency or availability issue in the inference phase, while ~\cite{liu2020securea,feng2020multiparticipant,feng2022semisupervisedb,feng2022semisupervisedc} proposed to transfer the knowledge from the active party to help the passive party build a classifier. These works are out of the scope of this work. 





\begin{table*}[!ht]
	\caption{Main FL works employing SSL methods. }
    \centering
    \begin{tabular}{p{0.6cm}p{5.9cm}||p{1.8cm}|p{1.9cm}p{2.0cm}|p{3.0cm}}
	    \hline
		\multirow{2}{*}{Setting} & \multirow{2}{*}{Works} & \multicolumn{3}{c|}{Data setting} & \multirow {2}{*}{Usage of labeled data} \\
        \cline{3-5}
        ~ & ~ & \multicolumn{1}{c}{labeled} & \multicolumn{1}{c}{unlabeled} & ~ & ~ \\
	    \hline
	    \hline
        \multirow{2}{*}{HFL} & FedMOON\cite{li2021modelcontrastive}, Fed-PCL\cite{tan2022federatedb}, FedProc\cite{mu2021fedproc} & \multicolumn{1}{c}{$\surd$} & ~ & ~ &  \scriptsize used in end-to-end training \\
        ~ & FedCA\cite{zhang2020federated}, FedU\cite{zhuang2021collaborative}, FedEMA\cite{zhuang2022divergenceaware}, FedX\cite{han2022fedxa} & \multicolumn{1}{c}{$\surd$} & \multicolumn{1}{c}{$\surd$} & ~ & \scriptsize used in finetuning  \\
        \hline
        ~ & ~ & \multicolumn{1}{c}{aligned labeled} & \multicolumn{1}{c}{aligned unlabeled} & \multicolumn{1}{c}{unaligned unlabeled} & ~ \\
        \hline
		\multirow{4}{*}{VFL} & FedCVT\cite{kangyan2022fedcvt}, FedMC\cite{yang2022multiview} & \multicolumn{1}{c}{$\surd$} & ~ & \multicolumn{1}{c|}{$\surd$} & \scriptsize used in end-to-end training \\
		~ & VFed-SSD\cite{li2022achieving} & \multicolumn{1}{c}{$\surd$} & \multicolumn{1}{c}{$\surd$} & \multicolumn{1}{c|}{~} & \scriptsize used in finetuning \\
		~ & SS-VFL\cite{castiglia2022lssl}, VFLFS\cite{feng2022vertical} & \multicolumn{1}{c}{$\surd$} & ~ & \multicolumn{1}{c|}{$\surd$} &  \scriptsize used in finetuning \\
        \cline{2-6}
        
        ~ & FedHSSL(ours) & \multicolumn{1}{c}{$\surd$} & \multicolumn{1}{c}{$\surd$} & \multicolumn{1}{c|}{$\surd$} & \scriptsize used in finetuning \\
		\hline
	\end{tabular}
\label{table:SSL_works}
\end{table*}
\subsection{Privacy Attacks and Protections in VFL} VFL involves two kinds of privacy leakage: feature leakage and label leakage.~\cite{he2019mi} proposed model inversion attack to infer features of the passive party. However, in the practical VFL setting, parties typically have black-box knowledge on the model information of each other. Thus, it is challenging for the attacker to infer features of other parties. The literature has proposed two forms of label inference attacks in VFL: the gradient-based~\cite{oscar2022split} and the model-based~\cite{fu2022label}. The former often applies to binary classification, and the latter is difficult to be defended against, but it requires auxiliary training data.~\cite{oscar2022split} proposed three protection methods against gradient-based attacks. \cite{yang2020defending} proposed a data encoding protection mechanism called CoAE, which can thwart model-based attacks effectively in some scenarios. Cryptography-based protections are seldom applied to VFL that involves deep neural networks (DNN) for their high communication and computational cost. ~\cite{kang2021privacy} proposed a HE-protected interactive layer that protects the outputs of parties' local DNN without protecting gradients. Thus, it can not defend against label inference attacks.

\section{Preliminaries}
We review the concepts of vertical federated learning and self-supervised learning methods we adopt in this work.

\subsection{Vertical Federated Learning}\label{sec:vfl}
Vertical federated learning deals with scenarios where participating parties share the same set of samples but each holds a distinct portion of features of these samples. More specifically, often one party holds labels but may or may not owns features. This party is called \textit{active party} because it typically is the initiator of VFL training and inferencing, while other parties hold only features and are called \textit{passive parties}~\cite{liu2022vertical}. 

\begin{figure}[ht!]
	\centering
\includegraphics[width=0.33\textwidth] {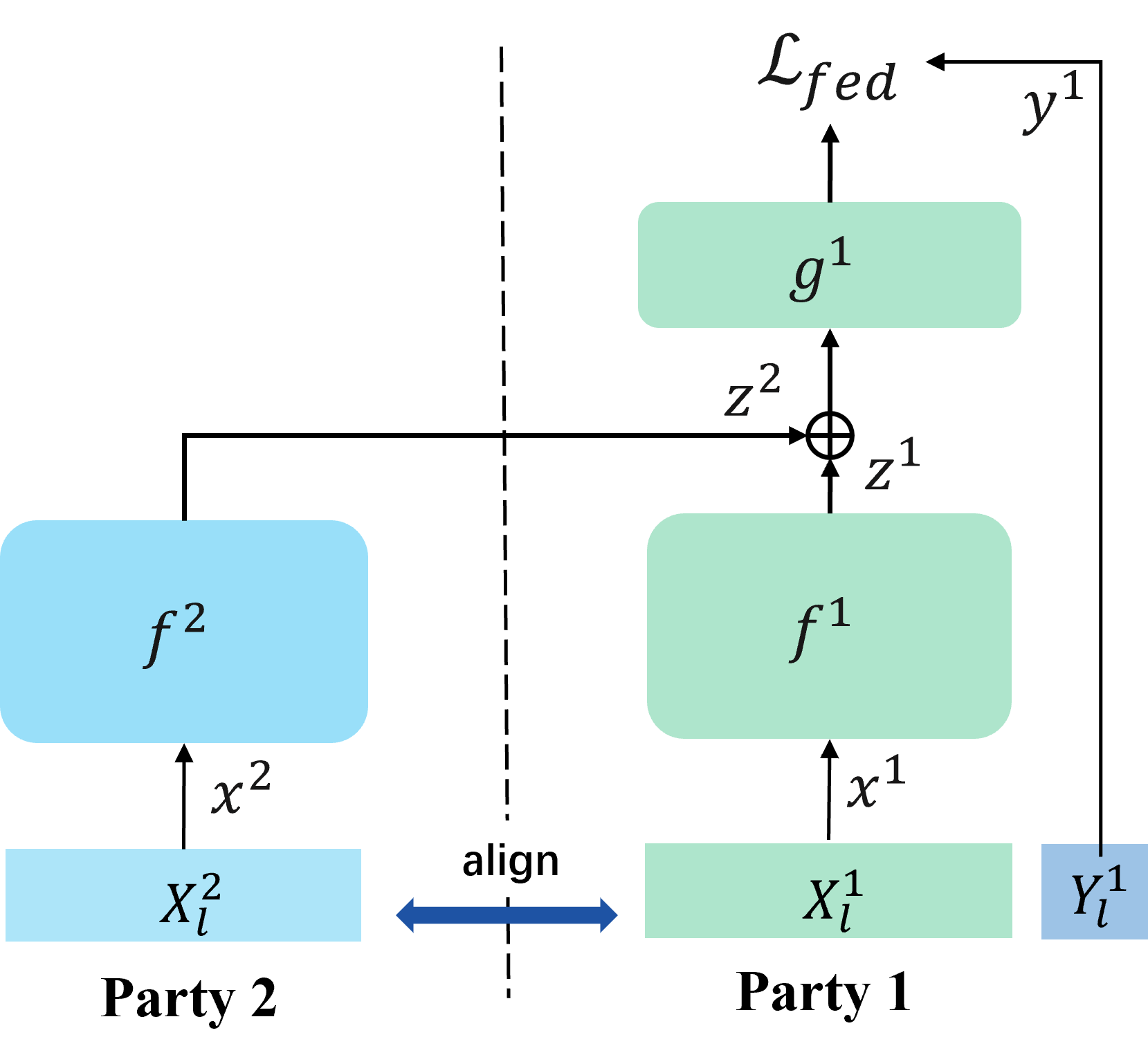}
	\caption{The conventional VFL setting illustrated by two parties. Active party 1 owns a bottom model $f^1$ and a top model $g^1$, while passive party 2 owns a bottom model $f^2$. We call the joint VFL model composed of $f^1$, $f^2$, and $g^1$ \textbf{FedSplitNN}.} 
	\label{fig:vfl_setting}
\end{figure}

We take the 2-party VFL setting as an example (see Figure \ref{fig:vfl_setting}). We assume the two parties collaboratively own a dataset $(Y^{1}_l, X^{1}_{l}, X^{2}_{l})$, party 1 is the active party who owns features and labels $(X^1_l, Y^1_{l})$, and party 2 is the passive party who owns features $X^2_l$. In this work, we use \textit{superscripts to identify participating party}, and subscripts for other denotations.


The active party 1 and passive party 2 utilize bottom model $f^1$ and $f^2$, respectively, to extracts high-level features from raw input $x^1 \in X^1_l$ and $x^2 \in X^2_l$. The active party also has a top model $g^1$ that transforms the aggregated (denoted by $\oplus$) outputs $z^1=f^1(x^1)$ and $z^2=f^2(x^2)$ into predicted labels, which
together with the ground-truth labels $y^1$ are used to compute the loss formulated in Eq.(\ref{eq:vfl_loss}). We call the joint VFL model composed of $f^1$, $f^2$, and $g^1$ \textbf{FedSplitNN}.
\begin{equation}
\label{eq:vfl_loss}
\begin{aligned}
    \mathcal{L}_{fed} = \ell_{ce}(g^1(z^1 \oplus z^2), y^1)
\end{aligned}
\end{equation}
where $\ell_{ce}$ is cross entropy, $y^1 \in Y^1_l$. Typical aggregation methods include concatenation along the feature axis, max-pooling and averaging. By minimizing $\mathcal{L}_{fed}$ in Eq. (\ref{eq:vfl_loss}), bottom model $f^1$ and $f^2$, as well as top model $g^1$ are updated.

\begin{figure*}[ht!]
	\centering
\includegraphics[width=0.85\textwidth] {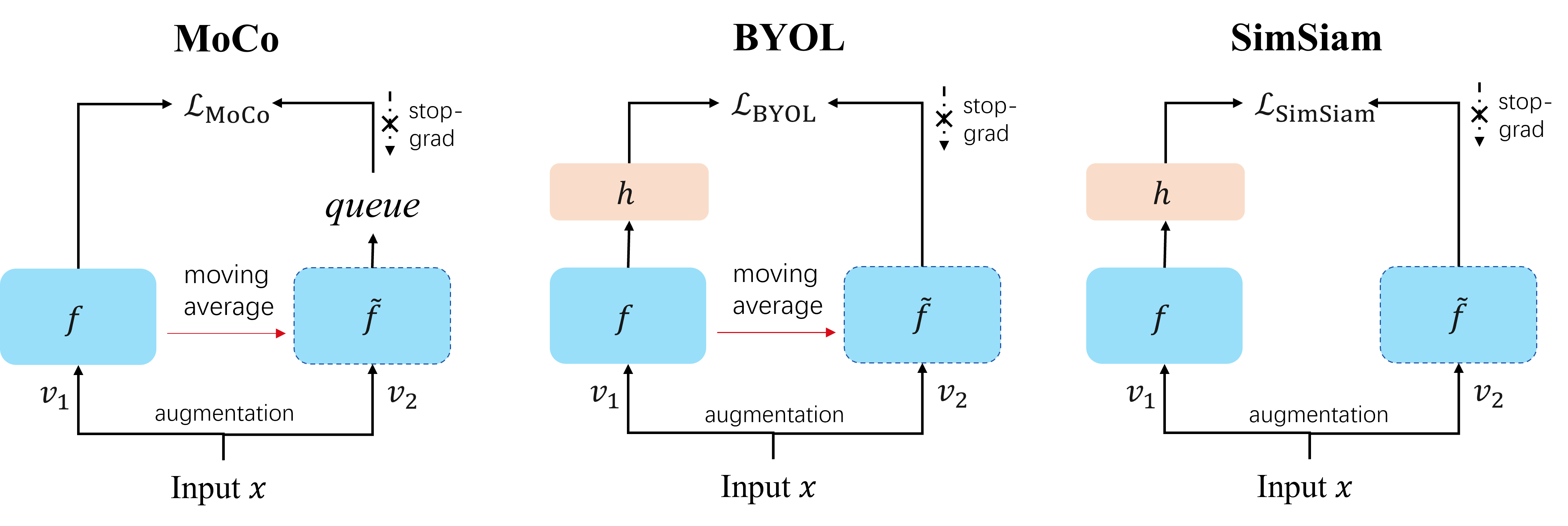}
	\caption{Architecture overview of three representative SSL methods. All methods comprise two encoders: an online encoder $f$, and a target encoder $\tilde{f}$. Gradients are not computed for the target encoder. For MoCo and BYOL, $\tilde{f}$ is the moving average of $f$. MoCo has a queue to provide additional negative samples for calculating InfoNCE loss. BYOL and SimSiam has a predictor on the top of online encoder, and use only positive pairs. For SimSiam, $\tilde{f} = f$. } 
	\label{fig:ssl_intro}
\end{figure*}


\subsection{Self-supervised learning}
Among various self-supervised learning (SSL) methods, contrastive learning~\cite{bachman2019learninga} has become the state-of-the-art method. It essentially groups semantically nearby samples (positive pairs) in the representation space while pushing apart the dissimilar samples (negative pairs) as far as possible~\cite{chen2020simple,he2020momentum}.~\cite{grill2020bootstrap,chen2021exploring} proposed non-contrastive methods, which use only positive pairs in self-supervised learning and demonstrates competitive performance with reduced complexity. 

\begin{table}[!h]
	\caption{Variations on the implementation of Algo.~\ref{alg:FedHSSL} for different SSL methods. MLP: multiple layer perceptron, EMA: exponential moving average.}
	\centering
	\begin{tabular}{l||c|c|c}
	        \hline
            \\[-1em]
		 \shortstack{ Method } & \shortstack{Target encoder $\tilde{f}$} & \shortstack{Predictor ($h$)}  & Loss\\
         \hline
         \hline
          \\[-1em]
        SimSiam &  equals online encoder $f$  &MLP & $\mathcal{L}_{\text{SimSiam}}$ \\
		\hline
          \\[-1em]
		BYOL  &  EMA of online encoder $f$ &  MLP   & $\mathcal{L}_{\text{BYOL}}$ \\
		\hline
       \\[-1em]
		MoCo  & EMA of online encoder $f$ & identical function & $\mathcal{L}_{\text{MoCo}}$ \\
     \hline
	\end{tabular}
 \label{table:ssl_intro}
\end{table}

In this section, we provide a brief introduction of three representative SSL methods: MoCo~\cite{he2020momentum}, BYOL~\cite{grill2020bootstrap}, SimSiam~\cite{chen2021exploring}. A schematic illustration of these three methods is shown in Fig. \ref{fig:ssl_intro}, and a comparison of their differences are listed in Table \ref{table:ssl_intro}. Given a batch of sample $x$, its two augmented version are $v_1=\mathcal{T}(x)$ and $v_2=\mathcal{T}(x)$. $\mathcal{T}$ denotes a data augmentation strategy. An online encoder $f$ transforms $v_1$ to $z_1$, and a \textit{target encoder} $\tilde{f}$ transforms $v_2$ to $\tilde{z}_2$. A predictor, $h$, is used to further convert $z_1$ to $p_1$. That is $z_1 = f(v_1)$, $\tilde{z}_2 = \tilde{f}(v_2)$, and $p_1 = h(z_1)$. All three methods follow this two-tower structure, and it should be noted that gradients are not computed for the target encoder. Here we omit the symmetrized computation path by swapping $v_1$ and $v_2$ for the simplicity.

\noindent\textbf{MoCo.}
Momentum Contrast (MoCo)~\cite{he2020momentum} utilizes the InfoNCE loss and a momentum encoder to ensure a better representation consistency and an additional queue to enable training with small batch size. That means $\tilde{f}$ is a momentum version of $f$, and a sample $Q$, which maintains a dynamic pool of feature vectors from previous batches. The predictor $h$ is simply an identical function. The training objective is
\begin{equation}
 \mathcal{L}_{\text{MoCo}} = - log \frac{exp(z_1 \cdot \text{st}(\tilde{z}_2))}{exp(z_1 \cdot \text{st}(\tilde{z}_2)) + 
 \sum_{\tilde{z}_q \in Q}exp(z_1 \cdot \text{st}(\tilde{z}_q))}
\label{eq:moco}
\end{equation}
where $\tilde{z_q} \in Q$, st(·) meas stop-gradient. By minimizing this loss, the positive pairs are pulled closer while negative pairs are pushed away in representation space.

\noindent\textbf{BYOL.}
Bootstrap Your Own Latent (BYOL)~\cite{grill2020bootstrap} differs from the MoCo method in that it only requires positive pairs, making the training procedure much simpler. The target encoder $\tilde{f}$ is a momentum version of $f$, the same as MoCo. To avoid a collapse in representation space, a multi-layer perceptron (MLP) is used as the predictor $h$. The training objective is formulated as follows. 
\begin{equation}
\begin{aligned}
 \mathcal{L}_{\text{BYOL}} & = \Vert \frac{p_1}{\|p_1\|_2} - \frac{\text{st}(\tilde{z}_2)}{\|\text{st}(\tilde{z}_2)\|_2} \Vert^2_2 \\
 & =  2 - 2 \cdot \frac{\langle p_1, \text{st}(\tilde{z}_2)\rangle}{\|p_1\|_2  \cdot \|\text{st}(\tilde{z}_2)\|_2}.
\label{eq:byol}
\end{aligned}
\end{equation}

\noindent\textbf{SimSiam.}
The Simple Siamese (SimSiam)~\cite{chen2021exploring} method is similar to BYOL that it also utilizes an asymmetric MLP predictor, $h$, and a similarity-based objective that only needs positive pairs. It further removes the momentum encoder and uses the same encoder, $\tilde{f} = f$, for converting $v_1$ and $v_2$. The training objective becomes:
\begin{equation}
 \mathcal{L}_{\text{SimSiam}} =  - \frac{p_1}{||p_1||_2}\cdot \frac{\text{st}(\tilde{z}_2)}{||\text{st}(\tilde{z}_2)||_2}.
\label{eq:simsiam}
\end{equation}

In this work, we adopt the three representative SSL methods,  SimSiam~\cite{chen2021exploring}, BYOL~\cite{grill2020bootstrap}, and MoCo~\cite{he2020momentum}, as the base SSL methods for FedHSSL to investigate the effectiveness of FedHSSL as a framework. 
 
\section{Methodology}
\label{sec:methods}
In this section, we formulate our VFL setting and problem. We then elaborate on our FedHSSL framework.

\begin{figure}[ht!]
	\centering
	\includegraphics[width=0.44\textwidth] {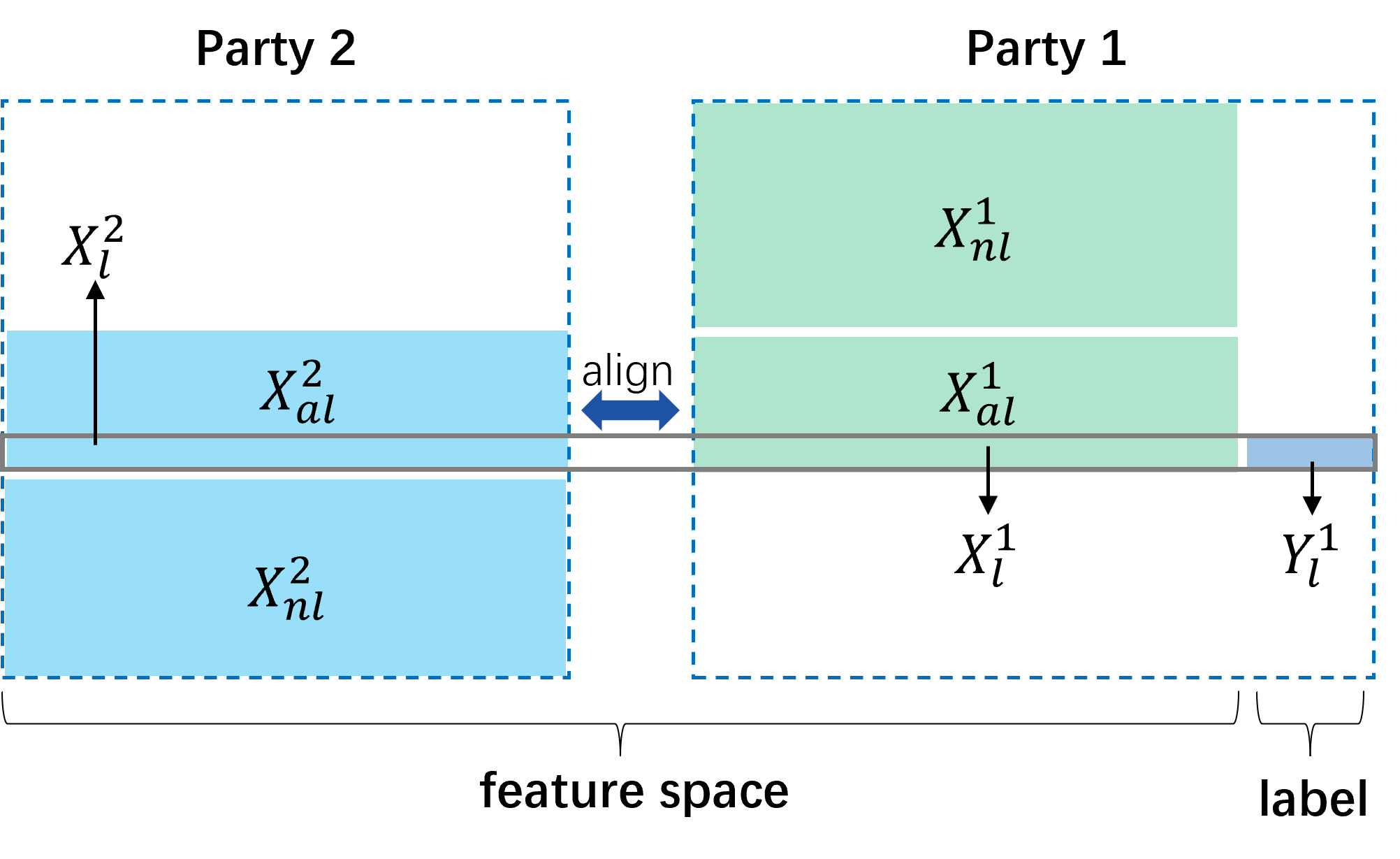}
	\caption{Virtual dataset owned by two parties. The aligned samples ($X^1_{al}, X^2_{al}$) account for a small portion of each party's total samples. The amount of labeled aligned samples ($Y^1_{l}, X^1_{l}, X^2_{l}$) is even less, while each party has a large amount of non-aligned local samples (i.e., $X^1_{nl}$ and $X^2_{nl}$).} 
	\label{fig:dataset_setting}
\end{figure}

\subsection{Problem Formulation}\label{sub_sec:problem}We consider a general VFL setting that involves $K$ parties. The $i_{th}$ party owns a dataset $X^i=({X}^i_{al}, {X}^{i}_{nl}),~i \in \{1,\dots,K\}$, where ${X}^i_{al}$ and ${X}^{i}_{nl}$ denote aligned and non-aligned samples, respectively. We assume only party $1$ has labels and denote party 1's labeled samples as $(Y^{1}_l, X^{1}_{l})$, where $X^{1}_{l} \subseteq {X}^1_{al}$. Figure \ref{fig:dataset_setting} depicts the virtual dataset formed by two parties (i.e., parties 1 and 2) for illustrative purposes.

In conventional VFL, as explained in Section \ref{sec:vfl}, participating parties collaboratively train a joint model only using aligned and labeled samples $(Y^{1}_l, X^{1}_{l}, X^{2}_{l},\dots, X^{K}_{l})$, leaving each party $i$'s aligned but unlabeled samples $X^i_{al} \backslash X^i_{l}$ as well as unaligned samples $X^i_{nl}$ unused. 

We propose a Federated Hybrid SSL (FedHSSL) framework that pretrains participants' local models by leveraging all available unlabeled samples of all parties  ${X}^i = ({X}^{i}_{al}, {X}^{i}_{nl})$ for $i, i \in \{1,\dots,K\}$. Then, the conventional VFL is conducted to fine-tune pretrained models with a classifier $g$ on top of pretrained models using aligned and labeled samples.

The goal of FedHSSL is to enhance the performance of the VFL joint model trained on downstream supervised task (see Section \ref{eq:vfl_loss}). Therefore, we evaluate the performance of FedHSSL on downstream supervised tasks.


\begin{figure*}[ht!]
	\centering
	\includegraphics[width=0.94\textwidth] {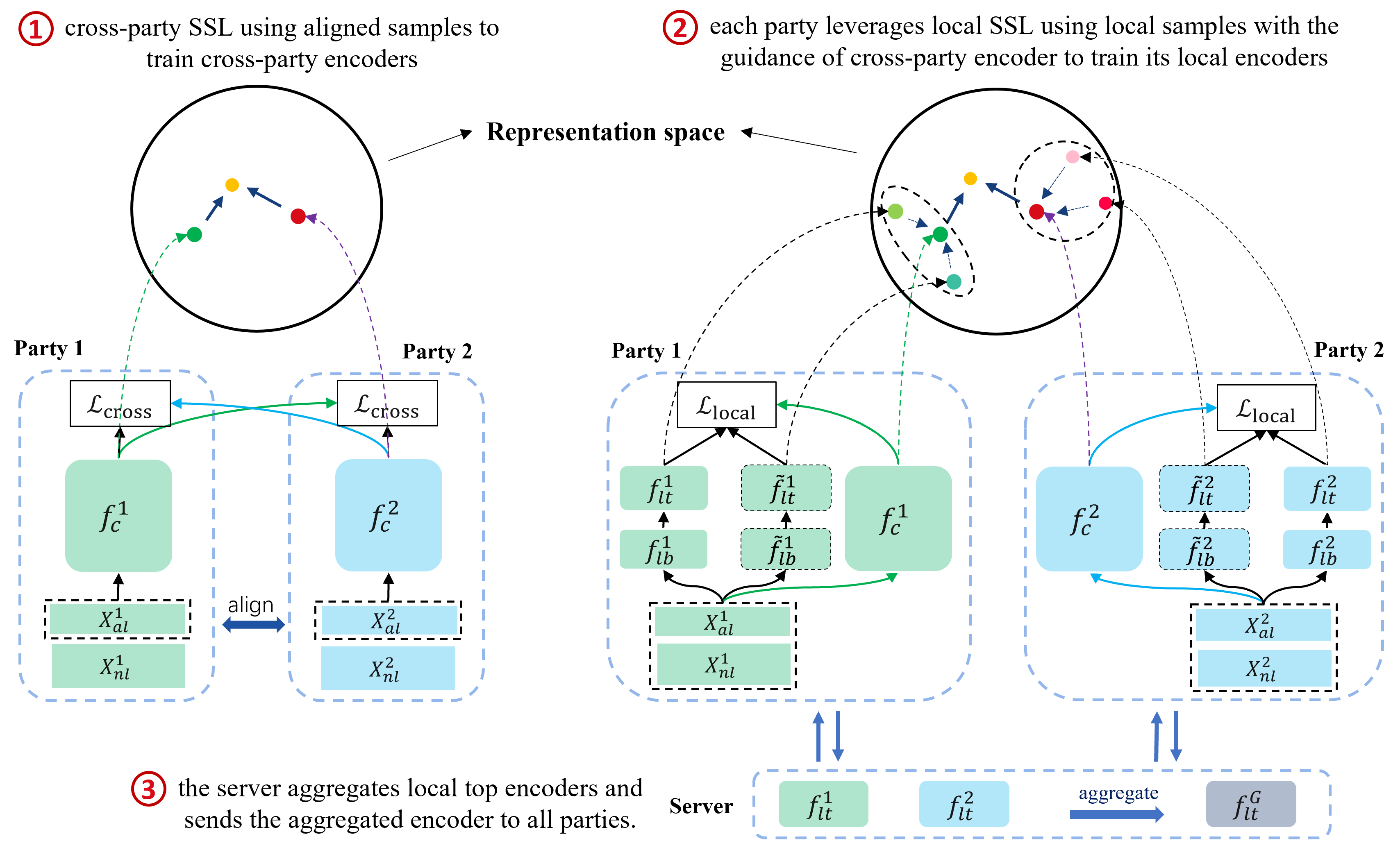}
	\caption{Overview of FedHSSL. Each party has a two-tower structured model. FedHSSL involves 3 steps: \numcircledmod{1} cross-party SSL using aligned samples to train cross-party encoders $f_c^1$ and $f_c^2$; \numcircledmod{2} each party $i$ leverages local SSL with the guidance of $f_c^i$ to train its local encoders $f_{lt}^i$ and $f_{lb}^i$ using local samples; \numcircledmod{3} the server aggregates local top encoders $f_{lt}^1$ and $f_{lt}^2$, and sends the aggregated encoder $f_{lt}^G$ to all parties. We omit predictors in this figure for brevity.}
	\label{fig:overview}
\end{figure*}


\begin{algorithm}[!ht] 
\caption{FedHSSL Pretraining Procedure}
\label{alg:FedHSSL}

\textbf{Input:} \\
Dataset ${X}^i = ({X}^{i}_{al}, {X}^{i}_{nl})$ of party $i, i \in \{1,\dots,K\}$;\\
Cross-party encoder $f^{i}_{c}$ and predictor $h^{i}_{c}, i \in \{1,\dots,K\}$;\\
Local encoder $f^{i}_{l}$=$(f^{i}_{lb}, f^{i}_{lt})$ and predictor $h^{i}_{l}, i \in \{1,\dots,K\}$;\\
\textbf{Output:} \\
Pretrained encoders $f^{i}_{c}$ and $f^{i}_{l}$, $i \in \{1,\dots,K\}$ \vskip 3pt
\begin{algorithmic}[1]


\STATE \gray{// \footnotesize Refer to Table \ref{table:ssl_intro} for implementation variations of adopting different SSL methods (i.e., SimSiam, BYOL, and MoCo)}

\FOR{each global iteration}

\STATE \gray{$\triangleright$ \footnotesize \textit{Step} \numcircledmod{1}: \textit{Cross-party SSL}}
\FOR {party $i \in \{1,\dots,K\}$}
    \FOR {mini-batch $x^i_{al} \in {X}^{i}_{al}$}
        \STATE Compute $z^{i}_c = f^{i}_{c}(x^i_{al})$ and $p^i_c = h^{i}_{c}(z^{i}_c)$
        \IF {$i == 1$} 
            \STATE Send $z^i_c$ to parties $\{2,\dots,K\}$;
        \ELSE
            \STATE Send $z^i_c$ to party $1$;
        \ENDIF
        \STATE Compute $\mathcal{L}^{i}_{\text{cross}}$ according to Eq. (\ref{eq:fedcssl})
        \STATE Update model $f^{i}_{c}$ and $h^{i}_{c}$
    \ENDFOR
\ENDFOR \vskip 3pt

\STATE \gray{$\triangleright$ \footnotesize \textit{Step} \numcircledmod{2}: \textit{Cross party-guided local SSL}}
\FOR { party $i \in \{1,\dots,K\}$}
    \FOR {mini-batch $x^i \in {X}^{i}$} 
        \STATE $v^i_1, v^i_2$ = $\mathcal{T}(x^i), \mathcal{T}(x^i)$
        \STATE $p^{i}_{1,l}, \tilde{z}^{i}_{2,l} = h^{i}_{l}(f^{i}_{l}(v^i_1)), \tilde{f}^{i}_{l}(v^i_2)$
        \STATE Compute $p^{i}_{2,l}$ and $\tilde{z}^{i}_{1,l}$ by swapping $v^i_1$ and $v^i_2$
        \STATE \gray{// \footnotesize  $z^{i}_{1,c}$ and $z^{i}_{2,c}$ are for cross-party regularization}
        \STATE $z^{i}_{1,c}, z^{i}_{2,c} = f^{i}_{c}(v^i_1), f^{i}_{c}(v^i_2)$ 
        \STATE Compute $\mathcal{L}^{i}_{\text{local}}$ according to Eq. (\ref{eg:fedgssl})
        \STATE Update model $f^{i}_{l}$ and $h^{i}_{l}$
    \ENDFOR
\ENDFOR \vskip 3pt

\STATE \gray{$\triangleright$ \footnotesize \textit{Step} \numcircledmod{3}: \textit{Partial model aggregation}}
\FOR { party $i \in \{1,\dots,K\}$}
        \STATE Send local model $f^{i}_{lt} \circ h^{i}_{l}$ to the server

\ENDFOR \vskip 2pt
    \STATE  The server performs $f^{G}_{lt} \circ h^{G}_{l} = \frac{1}{K}\sum_{i=1}^K f^{i}_{lt} \circ h^{i}_{l}$   
    \STATE The server sends $f^{G}_{lt} \circ h^{G}_{l}$ back to all parties

\ENDFOR 

\end{algorithmic}
\end{algorithm}

\subsection{Federated Hybrid Self-Supervised Learning}\label{sec:3-steps}



The core idea of FedHSSL is to utilize cross-party views (i.e., dispersed features) of samples aligned among parties and local views (i.e., augmentations) of samples within each party to improve the representation learning capability of the joint ML model through SSL. FedHSSL further utilizes generic features shared among parties to boost the joint model through partial model aggregation. Specifically, our FedHSSL consists of three steps:
\begin{enumerate}
\item Cross-party SSL using aligned samples;
\item Cross-party-guided local SSL using local samples;
\item Partial model aggregation. 
\end{enumerate}

These steps combine the VFL-like cross-party SSL and the HFL-like model aggregation, and thus we call them Federated Hybrid SSL (\textbf{FedHSSL}) as a whole. The training procedure of FedHSSL is described in Algo.~\ref{alg:FedHSSL} and illustrated in Fig. \ref{fig:overview}.

\subsubsection{Cross-Party SSL}
In VFL, each party can be thought of as holding one view of each aligned sample. These cross-party views naturally form positive sample pairs to train the SSL model (i.e., the cross-party encoder $f_c^i$ and predictor $h_c^i$) of each party $i$. The cross-party SSL is described in Step \numcircledmod{1} of Algo.~\ref{alg:FedHSSL}. Specifically, for each party $i$, its input $x^i$ is converted by the cross-party encoder $f^i_{c}$ to the representations $z^i_{c}$, which in turn is transformed to $p_c^i$ via a predictor $h_c^i$. Then, party 1 (with labels) exchanges its representations $z^1_{c}$ with other parties' representations $z^j_{c}, j = 2,\dots,K$. Upon receiving corresponding representations, each party $i$ optimize its cross-party model via minimizing the cross-party loss $\mathcal{L}^{i}_{\text{cross}}$:
\vspace{-0.4em}
\begin{equation}
\label{eq:fedcssl}
    \mathcal{L}_{\text{cross}}^i= \begin{cases}
        \frac{1}{K-1}\sum_{j=2}^{K} \mathcal{L}_{\text{SSL}}(p^1_{c}, z^j_{c}), & \text{if $i=1$}.\\[5pt]
        \mathcal{L}_{\text{SSL}}(p^i_{c}, z^1_{c}), & \text{otherwise}.
  \end{cases}
\end{equation}
where $\mathcal{L}_{\text{SSL}}$ is a self-supervised loss and its specific form depends on the specific SSL method applies to FedHSSL (see Table \ref{table:ssl_intro}). 

FedHSSL adopts the same message-exchanging strategy as the conventional VFL, in which messages are only exchanged between active party 1 and passive parties, mainly for communication efficiency. The difference is that FedHSSL exchanges no gradient between parties, which automatically implements the stop-gradient. 


\subsubsection{Cross-Party-Guided Local SSL}
We propose that each party $i$ uses its trained cross-party encoder $f_c^i$ as guidance to regularize its SSL training of local encoder $f_l^i$ and predictor $h_l^i$ using its local samples. The knowledge from the cross-party encoder helps improve the discriminative capability of $f_l^i$ and $h_l^i$. Besides, it encourages the representations generated by local encoders of different parties to be aligned in the representation space, which is beneficial for the partial model aggregation (i.e., the next step).


The cross-party-guided local SSL is described in Step \numcircledmod{2} of Algo.~\ref{alg:FedHSSL}. More specifically, for each party $i$, two randomly augmented views $v^i_1 = \mathcal{T}(x^i)$ and $v^i_2=\mathcal{T}(x^i)$ of an input $x^i$ are converted by a local online encoder $f^i_{l}$ and a local target encoder $\tilde{f}^i_{l}$ to the representations $z^i_{1,l}$ and $\tilde{z}^i_{2,l}$, respectively. $\mathcal{T}$ denotes a data augmentation strategy. A local predictor $h^i_{l}$ then transforms $z^i_{1,l}$ to $p^i_{1,l}$. 
Following~\cite{grill2020bootstrap}, we swap $v^i_1$ and $v^i_2$ to obtain $p^i_{2,l}$ and $\tilde{z}^i_{1,l}$. 
Then, party $i$ conducts the local SSL by minimizing the symmetrized loss:
\vspace{-0.4em}
\begin{equation}
\label{eg:fedgssl}
\begin{aligned}
    \mathcal{L}^i_{\text{local}} = &\frac{1}{2} \left(\mathcal{L}_{\text{SSL}}(p^i_{1,l}, \tilde{z}^i_{2,l} ) + \mathcal{L}_{\text{SSL}}(p^i_{2,l}, \tilde{z}^i_{1,l})\right)~+ \\ 
    & \gamma \left( \mathcal{L}_{\text{SSL}}(p^i_{1,l}, z^i_{1,c}) + \mathcal{L}_{\text{SSL}}(p^i_{2,l}, z^i_{2,c})\right),
\end{aligned}
\end{equation}
where $\mathcal{L}_{\text{SSL}}(p^i_{1,l}, z^i_{1,c}) + \mathcal{L}_{\text{SSL}}(p^i_{2,l}, z^i_{2,c})$ is the regularization imposed by the cross-party encoder $f^i_{c}$ on the training of local encoder; $z^i_{1,c} = f^i_{c}(v^i_1)$ and $z^i_{2,c}=f^i_{c}(v^i_2)$; $\gamma$ controls the strength of the regularization.  

The effect of cross-party guidance can be visualized in the representation space illustrated in Step \numcircledmod{2} of Figure \ref{fig:overview}: representations independently learned by the local SSL of each party tend to disperse to different locations in the representation space; with the guidance of the cross-party encoder, they are forced towards the position of cross-party encoders, which are trained to share similar behaviors in Step \numcircledmod{1}.


\subsubsection{Partial Model Aggregation (PMA)} An effective model aggregation requires that the models to be aggregated have sufficiently similar parameter distribution. The cross-party guided local SSL (Step \numcircledmod{2}) encourages the local encoders and their corresponding predictors $f_l^i \circ h_l^i,~i \in \{1,\dots,K\}$ to learn similar feature projection in the representation space, making $f_l^i \circ h_l^i,~i \in \{1,\dots,K\}$ potential candidates for partial model aggregation. 

We further divide the local encoder  $f_{l}^i$ of each party $i$ into a party-specific \textit{local bottom} encoder $f_{lb}^i$ and a \textit{local top} encoder $f_{lt}^i$, and share $f_{lt}^i \circ h_l^i $ with the server for aggregation. The rationale behind this design choice is two-fold: First, the local top encoder tends to learn a generic set of features, making it suitable to be shared among parties. Second, keeping the local bottom encoder private is beneficial for preventing parties' input features from being attacked (e.g., gradient inversion attack) by the server~\cite{yan2021fedcg}. The model aggregation is described in Step \numcircledmod{3} of Algo.~\ref{alg:FedHSSL}.



\textbf{Implementation Variations for Different SSL Methods.} 
We integrate \textbf{SimSiam}~\cite{chen2021exploring}, \textbf{BYOL}~\cite{grill2020bootstrap}, and \textbf{MoCo}~\cite{he2020momentum}, respectively, into FedHSSL to investigate the effectiveness of FedHSSL as a framework. The three SSL methods have three design differences leading to variations in the implementation of Algo.~\ref{alg:FedHSSL}, which are summarized in Table \ref{table:ssl_intro}.




\section{Experiments}\label{sec:experiments}


\subsection{Experimental Setup}

In this section, we elaborate on the experimental setup, including datasets, models, baselines, and training details.

\textbf{Datasets \& models.} We conduct experiments on 4 datasets: NUSWIDE~\cite{nus-wide-civr09}, Avazu~\cite{wang2014clickthrough}, BHI~\cite{mooneybreast}, and Modelnet~\cite{wu20153d}. The former 2 are tabular datasets, while the latter 2 are image datasets. For NUSWIDE, Avazu, and BHI, we split features of the same samples into 2 parts to simulate \textit{2-party} VFL scenario. For Modelnet, we divide samples describing the same objects into 4 groups to simulate \textit{4-party} VFL scenario. Table \ref{table:models} shows chosen models corresponding to each dataset for all parties. All predictors consist of two fully-connected layers (FC). (see Appendix \ref{app_datasets} for more detail on datasets)

\begin{table}[!ht]
	\caption{Models for evaluation. Emb: embedding layer.}
	\centering
	\begin{tabular}{l||c|c}
	        \hline
		 Dataset & \shortstack{local and cross-party \\ encoders ($f_{l}$ and $f_{c}$)} & \shortstack{local top encoder \\ for PMA ($f_{lt}$)}    \\
         \hline
         \hline
          \\[-1em]
            NUSWIDE &  2 FC &  top 1 layer of $f_l$  \\
		\hline
		Avazu &  1 Emb + 2 FC  &  top 1 layer of $f_l$   \\
		\hline
		BHI & ResNet-18 & top three blocks of $f_l$ \\
		\hline
		Modelnet & ResNet-18 &  top three blocks of $f_l$ \\
     \hline
	\end{tabular}
\label{table:models}
\end{table}

\textbf{Training Details for FedHSSL.} 
In addition to using all local samples for local SSL, we experiment with $40\%$ aligned samples of a dataset to pretrain cross-party encoder and predictor (i.e., cross-party SSL) of FedHSSL. We show our experiment with $20\%$ aligned samples for pretraining in Appendix \ref{app_more_data}. $\gamma$ is set to $0.5$ for all datasets (we investigate the sensitivity of $\gamma$ in Appendix \ref{app_reg_lambda}). 

\textbf{Baselines.} To evaluate the performance of FedHSSL, we adopt multiple baselines that cover the VFL methods we surveyed in Section \ref{sec:rel_ssl} (see Table \ref{table:SSL_works}). 
\begin{itemize}
    
\item \textbf{Supervised.} The first two baselines are \textbf{LightGBM} (LGB) \cite{ke2017lightgbm} and \textbf{FedSplitNN} (see Figure \ref{fig:vfl_setting}), which are widely used supervised VFL models trained on \textit{labeled and aligned samples}.

\item \textbf{Semi-supervised.} We adopt \textbf{FedCVT}~\cite{kangyan2022fedcvt} as another baseline. FedCVT leverages \textit{labeled aligned and local unaligned samples} to train a joint model consisting of participating parties' local encoders and a global classifier. FedCVT only works on the 2-party scenario. 

\item \textbf{Self-supervised using local data.} We implement three baselines leveraging representative SSL methods, SimSiam, BYOL, and MoCo, respectively, to pretrain participating parties' local encoders and predictors using only \textit{local samples}. We name them \textbf{FedLocalSimSiam}, \textbf{FedLocalBYOL}, and \textbf{FedLocalMoCo}, respectively. The three baselines cover methods used in SS-VFL~\cite{castiglia2022lssl} and VFLFS~\cite{feng2022vertical}.


\item \textbf{Self-supervised using aligned data.} 
VFed-SSD~\cite{li2022achieving} pretrains participating parties' local encoders and predictors using only \textit{aligned unlabeled samples}, which is covered by \textbf{FedCSSL}, a sub-procedure of FedHSSL.

\end{itemize}



\begin{table*}[!ht]
 	\centering
	\caption{Performance comparison of FedHSSL (integrated with SimSiam, BYOL, MoCo, respectively) and baselines with a varying number of labeled samples. Top-1 accuracy is used as the metric for NUSWIDE and Modelnet, while AUC and F1-score are metrics for Avazu and BHI, respectively. \% of labeled and aligned samples applies only to FedHSSL.}
	\begin{tabular}{c|l||c|c|c|c|c|c}
	    \hline
		\multicolumn{2}{l||}{\# of labeled and aligned samples:} & 200 & 400 & 600 & 800 & 1000 & \# of parties\\
	    \hline
	    \hline
	\multirow{9}{*}{\shortstack{ NUSWIDE \\ (Top1-Acc)}} & LGB & 0.425 $\pm$ 0.015 & 0.465 $\pm$ 0.028 & 0.526 $\pm$ 0.012 & 0.556 $\pm$ 0.013 & 0.587 $\pm$ 0.012 & \multirow{9}{*}{2} \\
		& FedSplitNN & 0.495 $\pm$ 0.022 & 0.535 $\pm$ 0.027 & 0.560 $\pm$ 0.015 & 0.573 $\pm$ 0.014 & 0.591 $\pm$ 0.013  \\
		& FedCVT & 0.522 $\pm$ 0.019 & 0.555 $\pm$ 0.013 & 0.602 $\pm$ 0.003 & 0.621 $\pm$ 0.006 & 0.629 $\pm$ 0.014  \\
		& FedLocalSimSiam  & 0.505 $\pm$ 0.027 & 0.536 $\pm$ 0.018 & 0.596 $\pm$ 0.013 & 0.603 $\pm$ 0.019 & 0.612 $\pm$ 0.017 \\
    	& FedLocalBYOL & 0.514 $\pm$ 0.032 & 0.527 $\pm$ 0.029 & 0.585 $\pm$ 0.022 & 0.599 $\pm$ 0.028 & 0.606 $\pm$ 0.027 \\
		& FedLocalMoCo & 0.566 $\pm$ 0.033 & 0.596 $\pm$ 0.022 & 0.625 $\pm$ 0.017 & 0.634 $\pm$ 0.017 & 0.639 $\pm$ 0.019 \\
         \cline{2-7}
            \\[-1em]
        & FedHSSL-SimSiam & 0.607 $\pm$ 0.003 & 0.641 $\pm$ 0.008 & 0.651 $\pm$ 0.006 & 0.662 $\pm$ 0.006 & 0.670 $\pm$ 0.003  \\
		& FedHSSL-BYOL & 0.598 $\pm$ 0.025 & 0.624 $\pm$ 0.034 & 0.645 $\pm$ 0.012 & 0.659 $\pm$ 0.007 & 0.664 $\pm$ 0.004 \\
		& FedHSSL-MoCo & \textbf{0.615} $\pm$ 0.021 & \textbf{0.642} $\pm$ 0.012 & \textbf{0.658} $\pm$ 0.003 & \textbf{0.668} $\pm$ 0.005 & \textbf{0.670} $\pm$ 0.006 \\
		\hline
		\hline
       \multirow{10}{*}{\shortstack{Avazu \\ (AUC)}}  & LGB & 0.563 $\pm$ 0.016 & 0.568 $\pm$ 0.019& 0.595 $\pm$ 0.020& 0.621 $\pm$ 0.012& 0.620 $\pm$ 0.012 & \multirow{10}{*}{2}  \\
		& FedSplitNN & 0.588 $\pm$ 0.031 & 0.581 $\pm$ 0.013 & 0.599 $\pm$ 0.019 & 0.595 $\pm$ 0.008 & 0.615 $\pm$ 0.006   \\
		& FedCVT & 0.594 $\pm$ 0.026 & 0.606 $\pm$ 0.022 & 0.608 $\pm$ 0.029 & 0.637 $\pm$ 0.015 & 0.647 $\pm$ 0.013  \\
		& FedLocalSimSiam & 0.575 $\pm$ 0.007 & 0.585 $\pm$ 0.020 & 0.591 $\pm$ 0.016 & 0.608 $\pm$ 0.026 & 0.629 $\pm$ 0.024  \\
    	& FedLocalBYOL & 0.560 $\pm$ 0.029 & 0.597 $\pm$ 0.015 & 0.600 $\pm$ 0.024 & 0.601 $\pm$ 0.004 & 0.605 $\pm$ 0.013 \\
		& FedLocalMoCo & 0.573 $\pm$ 0.024 & 0.591 $\pm$ 0.017 & 0.584 $\pm$ 0.027 & 0.596 $\pm$ 0.004 & 0.601 $\pm$ 0.011 \\
      \cline{2-7}
        \\[-1em]
      & FedHSSL-SimSiam & \textbf{0.623} $\pm$ 0.016 & \textbf{0.636} $\pm$ 0.026 & \textbf{0.649} $\pm$ 0.008 & \textbf{0.648} $\pm$ 0.014 & \textbf{0.663} $\pm$ 0.007  \\ 
		& FedHSSL-BYOL & 0.615 $\pm$ 0.031 & 0.634 $\pm$ 0.028 & 0.631 $\pm$ 0.016 & 0.630 $\pm$ 0.013 & 0.648 $\pm$ 0.010 \\
  	& FedHSSL-MoCo & 0.616 $\pm$ 0.014 & 0.632 $\pm$ 0.011 & 0.638 $\pm$ 0.017 & 0.641 $\pm$ 0.009 & 0.658 $\pm$ 0.007 \\
		\hline
		\hline
		\multirow{8}{*}{\shortstack{BHI \\ (F1-Score)}} & FedSplitNN  & 0.731 $\pm$ 0.003 & 0.738 $\pm$ 0.002 & 0.754 $\pm$ 0.002 & 0.752 $\pm$ 0.002 & 0.760 $\pm$ 0.005 & \multirow{8}{*}{2} \\
	    & FedCVT & 0.742 $\pm$ 0.013 & 0.747 $\pm$ 0.011 & 0.755 $\pm$ 0.007 & 0.758 $\pm$ 0.006 & 0.782 $\pm$ 0.003 \\
		& FedLocalSimSiam  & 0.760 $\pm$ 0.010 & 0.764 $\pm$ 0.006 & 0.788 $\pm$ 0.005 & 0.785 $\pm$ 0.004 & 0.798 $\pm$ 0.006  \\
    	& FedLocalBYOL & 0.760 $\pm$ 0.007 & 0.769 $\pm$ 0.008 & 0.781 $\pm$ 0.005 & 0.786 $\pm$ 0.005 & 0.796 $\pm$ 0.003 \\
		& FedLocalMoCo & 0.763 $\pm$ 0.003 & 0.771 $\pm$ 0.008 & 0.784 $\pm$ 0.012 & 0.793 $\pm$ 0.002 & 0.800 $\pm$ 0.008 \\
        \cline{2-7}
        \\[-1em]
		& FedHSSL-SimSiam & 0.805 $\pm$ 0.009 & 0.816 $\pm$ 0.006 & 0.822 $\pm$ 0.003 & 0.823 $\pm$ 0.002 & 0.830 $\pm$ 0.002 \\
		& FedHSSL-BYOL & 0.791 $\pm$ 0.011 & 0.806 $\pm$ 0.004 & 0.821 $\pm$ 0.002 & 0.822 $\pm$ 0.004 & 0.825 $\pm$ 0.003 \\
        & FedHSSL-MoCo & \textbf{0.806} $\pm$ 0.007 & \textbf{0.817} $\pm$ 0.002 & \textbf{0.822} $\pm$ 0.004 & \textbf{0.829} $\pm$ 0.004 & \textbf{0.831} $\pm$ 0.002 \\
		\hline
		\hline
		\multirow{8}{*}{\shortstack{Modelnet \\ (Top1-Acc)}} & FedSplitNN & 0.612 $\pm$ 0.019 & 0.684 $\pm$ 0.011 & 0.733 $\pm$ 0.002 & 0.765 $\pm$ 0.007 & 0.771 $\pm$ 0.005 & \multirow{8}{*}{4}  \\
		& FedLocalSimSiam & 0.622 $\pm$ 0.022 & 0.698 $\pm$ 0.017 & 0.761 $\pm$ 0.009 & 0.779 $\pm$ 0.004 & 0.797 $\pm$ 0.006  \\
  	    & FedLocalBYOL & 0.635 $\pm$ 0.004 & 0.707 $\pm$ 0.010 & 0.760 $\pm$ 0.007 & 0.775 $\pm$ 0.009 & 0.794 $\pm$ 0.007\\
		& FedLocalMoCo & 0.659 $\pm$ 0.022 & 0.722 $\pm$ 0.012 & 0.784 $\pm$ 0.008 & 0.798 $\pm$ 0.007 & 0.815 $\pm$ 0.007 \\
        \cline{2-7}
             \\[-1em]
		& FedHSSL-SimSiam & \textbf{0.707} $\pm$ 0.009 & \textbf{0.772} $\pm$ 0.006  & \textbf{0.806} $\pm$ 0.008 & \textbf{0.826} $\pm$ 0.007 & \textbf{0.833} $\pm$ 0.006 \\
		& FedHSSL-BYOL & 0.681 $\pm$ 0.005 & 0.752 $\pm$ 0.002 & 0.800 $\pm$ 0.008 & 0.807 $\pm$ 0.007 & 0.825 $\pm$ 0.009 \\
  	& FedHSSL-MoCo & 0.705 $\pm$ 0.016 & 0.764 $\pm$ 0.012 & 0.804 $\pm$ 0.006 & 0.822 $\pm$ 0.003 & 0.830 $\pm$ 0.007 \\

        \hline
	\end{tabular}
\label{table:main_results_v2}
\end{table*}

All baselines and FedHSSL use the same amount of labeled and aligned samples for training or fine-tuning. For each dataset, the local encoders of FedHSSL and baselines have the same model architecture.

We evaluate FedHSSL methods and SSL baselines by fine-tuning its pretrained encoders and a classifier on top with a varying number of labeled samples ranging from 200 to 1000. Results are reported as averages over 5 trials (see more training details in Appendix \ref{app_training_detail}).



\textbf{Data Augmentation.} For BHI and Modelnet, data are augmented following the setting described in~\cite{chen2021exploring}. For NUWISDE, $30\%$ features are distorted by replacing the original value with a random value as described in~\cite{bahri2021scarf}. For Avazu, the continuous features are treated the same way as those of the NUSWIDE, while the categorical features are replaced by extra untrained embedding vectors as described in ~\cite{yao2021selfsupervised}.

\begin{figure}[!hb]
	\centering
	\includegraphics[width=0.45\textwidth] {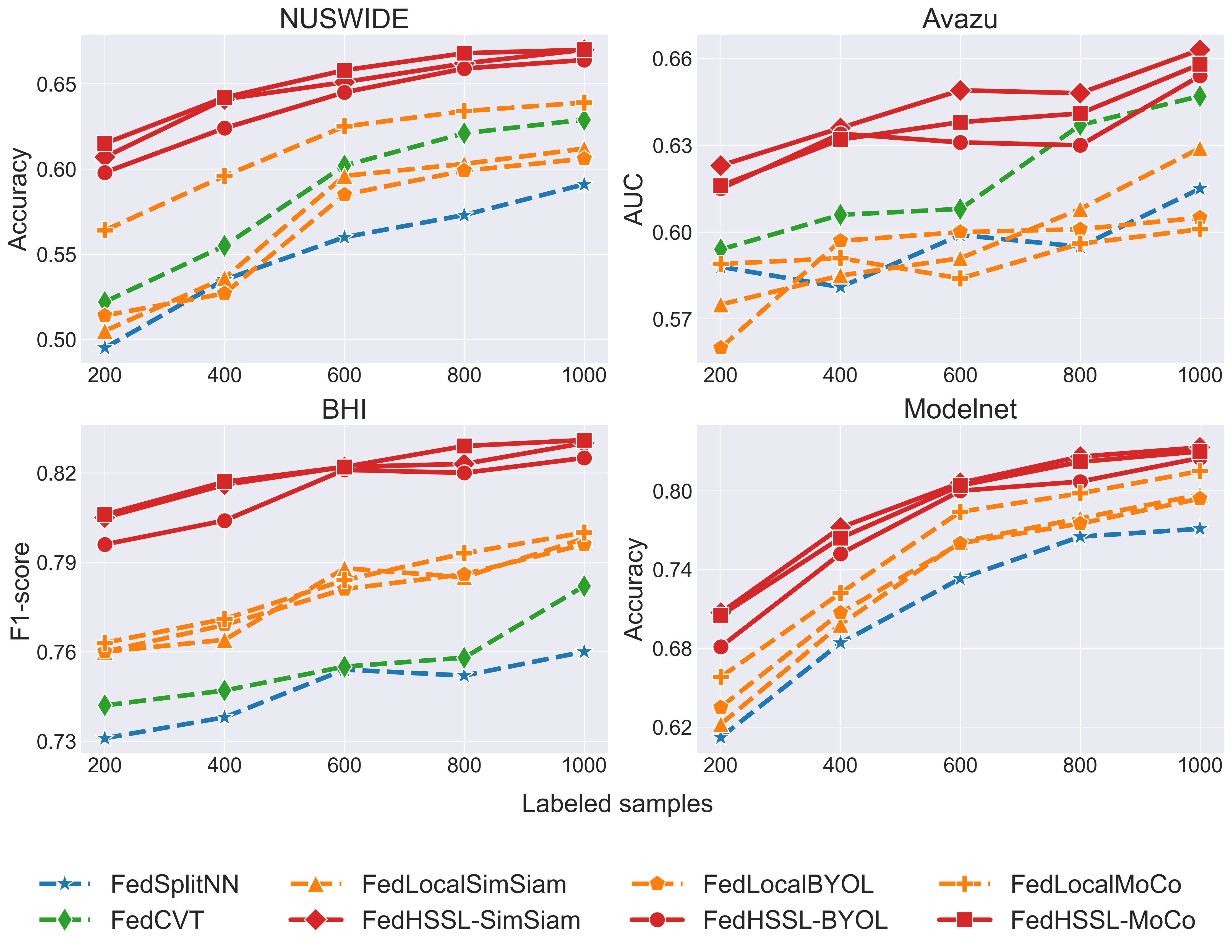}
 \vspace{-0.8em}
	\caption{Performance comparison of FedHSSL (integrated with SimSiam, BYOL, and MoCo, respectively) and baselines.}
	\label{fig:main_result_fedhssl}
\end{figure}

\subsection{Main Results}

We compare the performance of our FedHSSL framework integrated with SimSiam, BYOL, and MoCo, respectively, with the performance of baselines on four datasets. Both Table \ref{table:main_results_v2} and Figure \ref{fig:main_result_fedhssl} show the results. 

Figure \ref{fig:main_result_fedhssl} illustrates that FedHSSL methods (red) generally enhance performance compared with baselines by large margins for all datasets. For example, as reported in Table \ref{table:main_results_v2}, with 200 labeled samples, the performance of FedHSSL-SimSiam is improved by $0.102$ on NUSWIDE, by $0.048$ on Avazu, by $0.045$ on BHI and by $0.085$ on Modelnet, respectively, compared with FedLocalSimSiam. Similarly, FedHSSL-BYOL outperforms FedLocalBYOL by $0.084$, $0.055$, $0.031$, and $0.046$, respectively, on the 4 datasets; FedHSSL-MoCo outperforms FedLocalMoCo by $0.049$, $0.043$, $0.043$, and $0.046$, respectively, on the 4 datasets. Besides, with 200 labeled samples, the best-performing FedHSSL method outperforms FedCVT by $0.093$ on NUSWIDE, $0.029$ on Avazu, and $0.063$ on BHI, respectively. 

With more labeled samples involved in fine-tuning, the performance improvement of FedHSSL is still noticeable. For example, with 1000 labeled samples, the performance of FedHSSL-SimSiam is improved by $0.058$ on NUSWIDE, by $0.034$ on Avazu, by $0.032$ on BHI, and by $0.036$ on Modelnet, respectively, compared with FedLocalSimSiam.

\begin{figure}[!hb]
	\centering
	\includegraphics[width=0.48\textwidth] {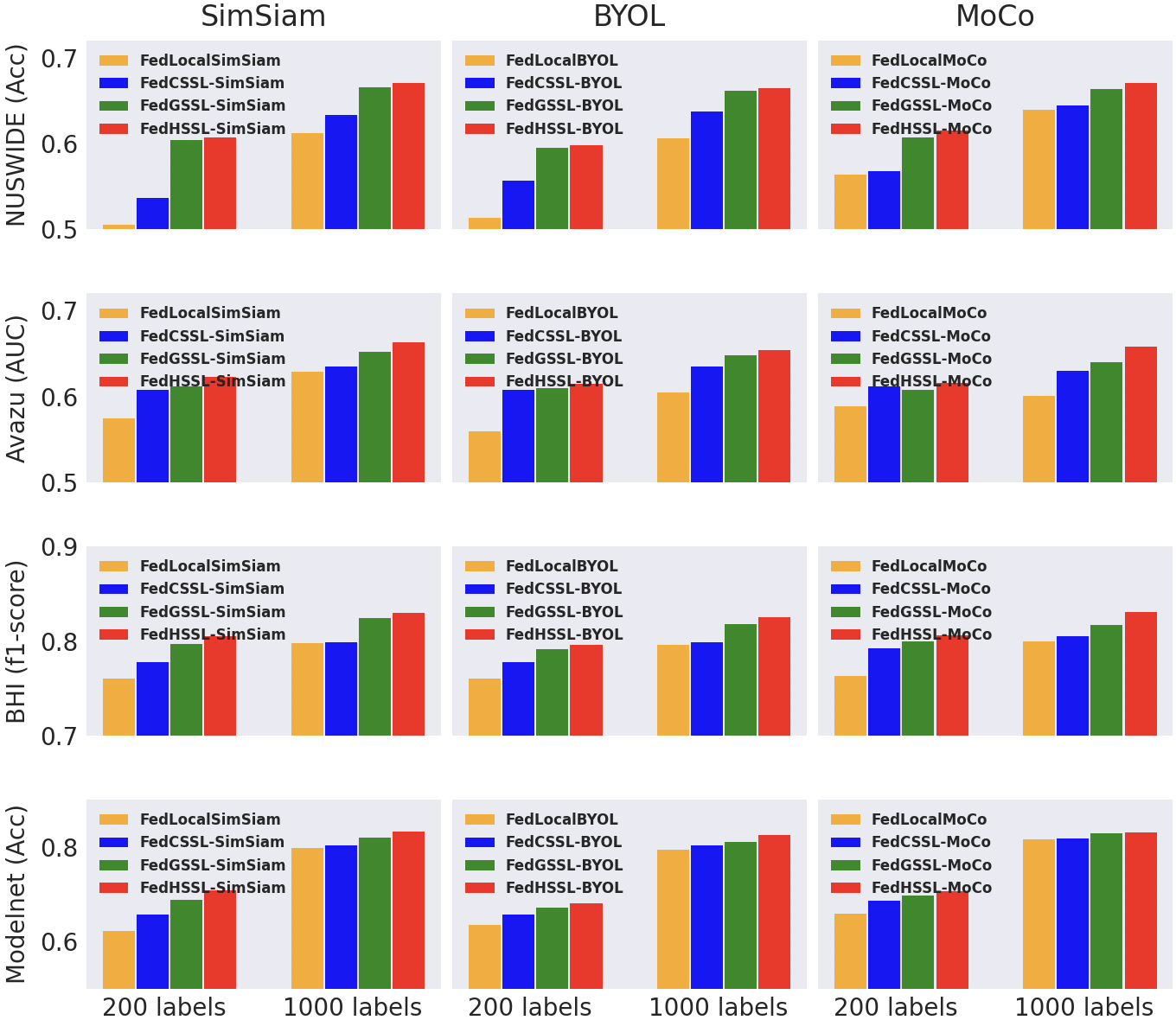}
\caption{Ablations on FedHSSL. We compare the performance of FedCSSL (blue), FedGSSL (green), and FedHSSL(red) for SimSam, BYOL, and MoCo, respectively. These methods are pretrained with all local samples and $40\%$ aligned samples and finetuned with a varying number of labeled and aligned samples. FedLocalSimSiam, FedLocalBYOL, and FedLocalMoCo are baselines for comparison.}
	\label{fig:ablations_fedhssl}
\end{figure}

\subsection{Ablation Study}\label{subseq:ablation}
To study the effectiveness of each step in FedHSSL, we consider two sub-procedures of FedHSSL: (i). \textbf{FedCSSL}, which is the cross-party SSL step in Algo.~\ref{alg:FedHSSL} (i.e., Step \numcircledmod{1}). (ii). \textbf{FedGSSL}, which is FedCSSL + cross-party-guided local SSL step in Algo.~\ref{alg:FedHSSL} (i.e., Step \numcircledmod{1} + Step \numcircledmod{2}). We evaluate FedCSSL and FedGSSL in the same way as that of FedHSSL: pretrained encoders are fine-tuned by minimizing Eq (\ref{eq:vfl_loss}) using aligned and labeled data.

\begin{table*}[ht]
 	\centering
	\caption{Study the impact of cross-party encoders on (1) local SSL and (2) partial model aggregation (PMA). The local encoders of FedLocalSimSiam, FedLocalBYOL, and FedLocalMoCo are pretrained using local SimSiam, BYOL, and MoCo, respectively. While the local encoders of $\text{FedGSSL-SimSiam}^*$, $\text{FedGSSL-BYOL}^*$, and $\text{FedGSSL-MoCo}^*$ are pretrained using \textit{cross-party-guided} SimSiam, BYOL, and MoCo, respectively. All methods are finetuned using 200 labeled samples. The down arrow $\red{\downarrow}$ indicates the performance decreases when the corresponding methods combine with PMA. The up arrow $\blue{\uparrow}$ indicates otherwise.}
	\begin{tabular}{l||c c|c c|c c|c c} 
	    \hline
		  & \multicolumn{2}{c|}{NUSWIDE} & \multicolumn{2}{c|}{Avazu} & \multicolumn{2}{c|}{BHI} & \multicolumn{2}{c}{Modelnet} \\
	    \cline{2-9}
          \\[-1em]
	   Method & $-$ & w/ PMA  & $-$ & w/ PMA & $-$ & w/ PMA &  $-$ & w/ PMA \\
	    \hline
		\hline
		FedLocalSimSiam & 0.505 & 0.537 $\blue{\uparrow}$ & 0.580 & 0.582 $\blue{\uparrow}$ & 0.760 & 0.743 $\red{\downarrow}$ & 0.622 & 0.599 $\red{\downarrow}$   \\
		$\text{FedGSSL-SimSiam}^*$ & 0.543 & 0.553 $\blue{\uparrow}$ & 0.606 & 0.609 $\blue{\uparrow}$  & 0.783 & 0.789 $\blue{\uparrow}$ & 0.679 & 0.688 $\blue{\uparrow}$ \\
		\hline
        FedLocalBYOL &  0.514 & 0.512 $\red{\downarrow}$ & 0.560 & 0.575 $\blue{\uparrow}$ & 0.760 & 0.756 $\red{\downarrow}$ & 0.635 & 0.629 $\red{\downarrow}$  \\
            $\text{FedGSSL-BYOL}^*$ & 0.543 & 0.544 $\blue{\uparrow}$ & 0.591 & 0.606 $\blue{\uparrow}$ & 0.778 & 0.785 $\blue{\uparrow}$ & 0.640 & 0.656 $\blue{\uparrow}$ \\
		\hline
        FedLocalMoCo & 0.566 & 0.563 $\red{\downarrow}$ & 0.573 & 0.587 $\blue{\uparrow}$ & 0.763 & 0.760 $\red{\downarrow}$ & 0.659 & 0.639 $\red{\downarrow}$ \\

      $\text{FedGSSL-MoCo}^*$ & 0.613 & 0.612 $\red{\downarrow}$ & 0.603 & 0.611 $\blue{\uparrow}$ & 0.787 & 0.795 $\blue{\uparrow}$ & 0.664 & 0.674  $\blue{\uparrow}$ \\
		\hline
	\end{tabular}
\label{table:impact}
\end{table*}

\textbf{The Effectiveness of Each Step Involved in FedHSSL.} 
Figure \ref{fig:ablations_fedhssl} illustrates that for each SSL method (i.e., SimSiam, BYOL, and MoCo on each column), FedCSSL consistently outperforms its corresponding FedLocalSSL as the number of labeled samples increases on the four datasets. By integrating local SSL into FedCSSL, FedGSSL generally enhances the performance over FedCSSL. The enhancement is significant on NUSWIDE (by $\approx0.05$ averagely) and noticeable on the other three datasets. By additionally conducting partial model aggregation (PMA), FedHSSL further boosts the performance on the four datasets. These results demonstrate the effectiveness of all three steps involved in FedHSSL.



\textbf{The Impact of Cross-Party Encoders' Guidance on Local SSL and Model Aggregation.} 
For a fair comparison, FedLocalSSL and $\text{FedGSSL}^*$ all use pretrained local encoders during fine-tuning. The star $*$ distinguishes $\text{FedGSSL}^*$ from $\text{FedGSSL}$, which leverages both cross-party and local encoders for fine-tuning. 

Table \ref{table:impact} reports that, for each SSL method (i.e., SimSiam, BYOL, and MoCo), $\text{FedGSSL}^*$ consistently outperforms its corresponding FedLocalSSL on all datasets. For example, FedGSSL-SimSiam outperforms FedLocalSimSiam by 0.038, 0.026, 0.023, and 0.057 on the four datasets, respectively. This demonstrates the effectiveness of the cross-party SSL in improving the representation learning of local SSL.

We further analyze the impact of cross-party encoders on partial model aggregation (PMA). Table \ref{table:impact} reports that directly combining FedLocalSSL and PMA may jeopardize the overall performance. For example, the performance of FedLocalSimSiam$+$PMA decreases by around $2\%$ compared with that of FedLocalSimSiam on BHI and Modelnet. Similar trends can be found on FedLocalBYOL$+$PMA and FedLocalMoCo$+$PMA. Assisted by the cross-party encoder, we observe a noticeable performance improvement on FedGSSL$^*+$PMA over FedGSSL$^*$ for all SSL methods generally across all datasets. This manifests that the guidance of cross-party encoders mitigates the heterogeneity among features of different parties so that it positively impacts PMA.

\subsection{Communication Efficiency}\label{subseq:comm}
The pretraining of FedHSSL utilizes all aligned samples, which results in higher communication overhead compared to conventional VFL that only uses labeled aligned samples. To mitigate this communication overhead, each party in FedHSSL can perform multiple updates in the cross-party SSL step (Step \numcircledmod{1} of Figure \ref{fig:overview}) to reduce communication rounds. Specifically, after received feature representations $z_c$ from other parties, each party conducts multiple local SSL updates by minimizing cross-party SSL loss (\ref{eq:fedcssl}) using $z_c$. This strategy is similar to FedBCD~\cite{liu2022fedbcd}, in which each party uses received gradients to update local model for multiple local updates. 

\begin{figure}[!ht]
	\centering
	\includegraphics[width=0.44\textwidth] {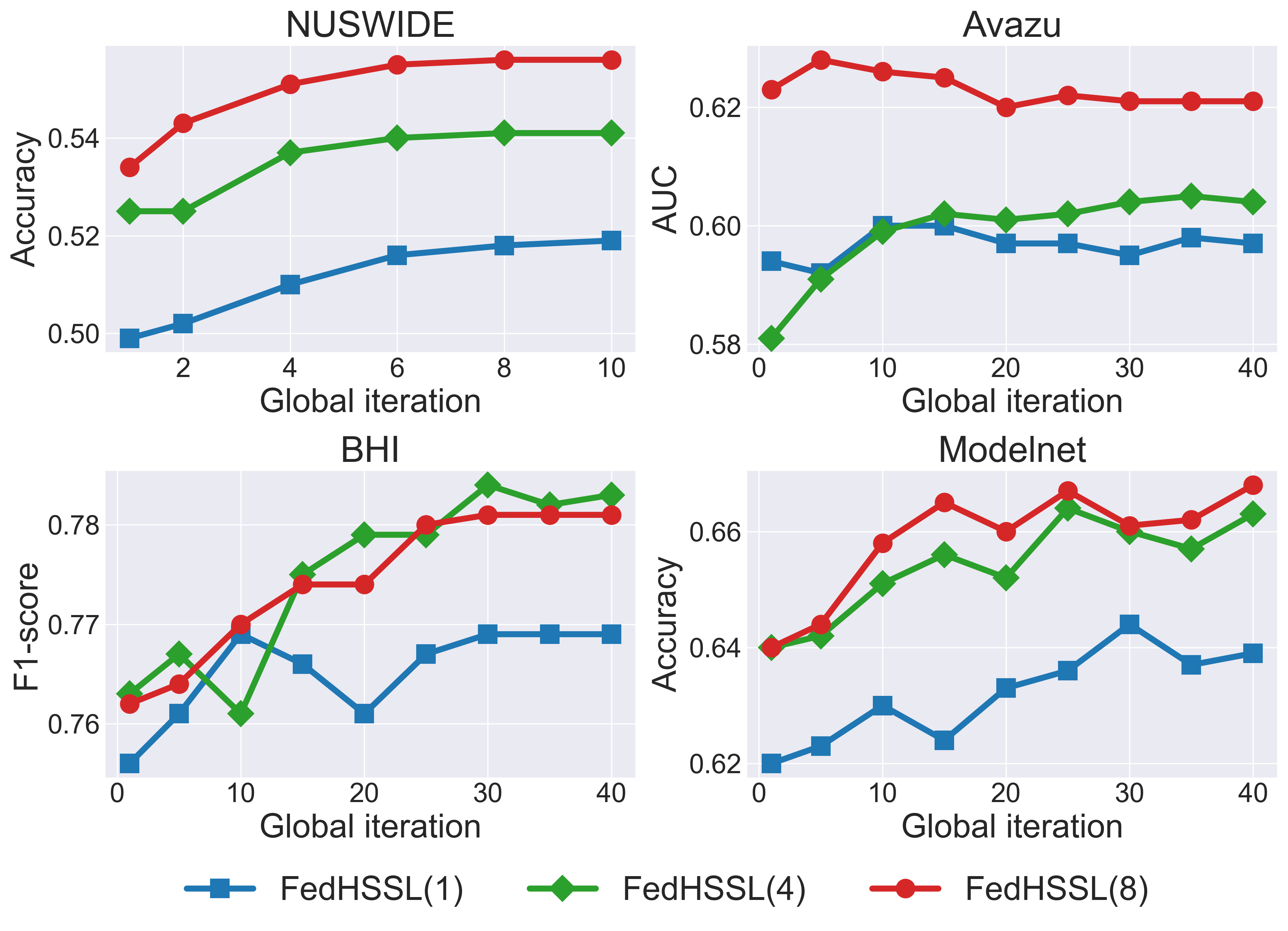}
	\vspace{-1em}
	\caption{Comparison of the performance of FedHSSL under different numbers of local updates in the cross-party SSL step. Results are obtained by averaging three rounds of experiments with different random seeds. $20\%$ training samples are aligned for cross-party SSL and 200 labeled samples are used in the fine-tuning. SimSiam is used as the default SSL method. }
	\label{fig:comm}
\end{figure}

We investigate the impact of multiple local updates in the cross-party SSL (Step \numcircledmod{1} of FedHSSL) on the communication efficiency by experimenting with various numbers of local updates in the range of ${1, 4, 8}$. We denote $e$ as the number of local updates. For these experiments, we adopt SimSiam as the base SSL method for FedHSSL. 

Figure \ref{fig:comm} illustrates the results. It shows that, with larger $e$, FedHSSL generally achieves better main task performance with the same global iterations on 4 datasets. However, on BHI, FedHSSL with 8 local updates performs worse than FedHSSL with 4 local updates, indicating that larger $e$ do not necessarily lead to better performance and an appropriate $e$ should be carefully chosen in order to achieve the best performance. 


\section{Privacy Analysis on Label Inference Attack}

In this section, we investigate whether FedHSSL, as a self-supervised VFL framework, can achieve a better privacy-utility trade-off against the label inference attack compared with baseline methods. We adopt SimSiam as the base SSL method for FedHSSL. Each party in FedHSSL pretrains its local model using all local samples and $20\%$ aligned samples. Supervised VFL training (including fine-tuning) is conducted using $200$ aligned and labeled samples.

\subsection{Threat Model} 
We first discuss the threat model, including the attacker's objective, capability, knowledge, and attacking methods.


\textbf{Adversary's objective}. We assume that party 2 is the  adversary who wants to infer labels $y^1$ owned by party 1.

According to the nature of dispersed data in our VFL setting, there can be three adversary objectives~\cite{kang2022framework}: (i) labels owned by the active party; (ii) features owned by the active party; (iii) features owned by the passive party. We focus on label inference attack where a passive party (i.e., party 2) is the adversary and it wants to infer labels $y^1$ owned by the active party (i.e., party 1) for the reasons that: (i) in the practical VFL setting, parties have black-box knowledge on the model information of each other, and thus it is highly challenging to party 1 to infer the features $x^2$ of party 2~\cite{he2019mi}; (ii) during model aggregation of FedHSSL, parties only share their local top encoders with the server while keeping the local bottom encoders private, in which case the server is not able to reconstruct features of any party~\cite{yan2021fedcg}; (iii) the labels owned by the active party is an important target for adversaries in VFL compared to HFL. Because in real-world VFL applications such as finance and advertisement, \textit{the labels may contain sensitive user information or are valuable assets}. 

\textbf{Adversary's capability}. We assume that the adversary party 2 is \textit{semi-honest} such that the adversary faithfully follows the vertical federated training protocol but it may mount privacy attacks to infer the private data of other parties.

\textbf{Adversary's knowledge}. In VFL, participating parties typically have blackbox knowledge about each other. However, adversaries may guess some of the knowledge about others according to the information they have. In this work, we assume that the information about the \textit{model structure}, \textit{input shape} and \textit{number of classes} supported by the active party's task is shared among parties. We also assume party 2 has a few \textit{auxiliary labeled samples} $\mathcal{D}_{aux}$

\subsection{Privacy attacking and protection mechanism}\label{sec:atta_prot}
\textbf{Privacy attacking mechanism}. There are mainly two kinds of label inference attacks in the VFL setting: the gradient-based attacks~\cite{oscar2022split} and the model-based attacks~\cite{fu2022label}. The former applies only to binary classification and can be thwarted effectively by state-of-the-art privacy protections (e.g., Marvell~\cite{oscar2022split}), while the latter is difficult to be prevented. In this work, we study the \textit{model completion} (MC) attack~\cite{fu2022label}, the representative of the model-based label inference attack. MC attack involves three steps: 
\begin{enumerate}
    \item Party 1 and party 2 conduct federated training, which can be FedHSSL pertaining or fine-tuning phase of downstream tasks. Upon the completion of training, party 2 obtains trained local models $f^2$; 
    \item Party 2 constructs a complete attacking model $\mathcal{A}_{\text{FedHSSL}}$ by training an inference head $g^2$ on top of $f^2$ using few auxiliary labeled data;
    \item Party 2 infers labels of its inference data $x_{\text{inf}}^2$ through $y^2_{\text{inf}}$ = $\mathcal{A}_{\text{FedHSSL}} (x^2_{\text{inf}})$ during inference phase.  
\end{enumerate}

Adversary party 2 can launch MC during the pretraining phase of FedHSSL or fine-tuning after FedHSSL. In this section, we study both scenarios. 
 
\textbf{Privacy protection mechanism}. we adopt \textit{isotropic Gaussian noise} (ISO)~\cite{oscar2022split} as the protection method. Specifically, party 1 perturbs model information $d \in \mathbb{R}^{b \times m}$ exposed to the adversary (i.e., party 2) by applying ISO to $d$, which can be forward embedding and backward gradients: 
\begin{equation}
\label{iso}
  \text{ISO}(d) = d + \varepsilon_{\text{iso}}
\end{equation}
where $\varepsilon_{\text{iso}} \sim \mathcal{N}(0, \sigma^2_{\text{iso}})$ is the noise added to protect privacy, $\sigma_{\text{iso}}=(\lambda \cdot ||d_{max}||_2) / \sqrt{m}$ is the standard deviation, and $||d_{max}||_2$ is the largest value in the batch-wise 2-norms $||d||_2$ of $d$, $\lambda$ is the noise amplifier and controls the strength of the ISO protection. We refer interesting readers to \cite{oscar2022split} for details on MC attack and ISO protection.
   
\textbf{Defending against Model Completion}
This experiment is conducted on FedHSSL-SimSiam pretraining. On the one hand, the adversary party 2 trains an attacking model $\mathcal{A}_{\text{FedHSSL}}$ according to the procedure described in Section \ref{sec:atta_prot}. On the other hand, party 1 applies ISO to the output of its cross-party encoder and parameters of its local top encoder to mitigate privacy leakage. After pretraining, party 2 leverages $\mathcal{A}_{\text{FedHSSL}}$ to predict labels of incoming samples.

For a fair comparison, we assume the adversary trains a baseline attacking model $\mathcal{A}_{\text{SimSiam}}$, pretrained by normal SimSiam, using $\mathcal{D}_{aux}$. Intuitively, $\mathcal{A}_{\text{SimSiam}}$ can be thought of as the adversary's prior knowledge on labels, while $\mathcal{A}_{\text{FedHSSL}}$ is the posterior knowledge on labels after the MC attacking.


\begin{table}[!h]
	\caption{Comparison of $\mathcal{A}_{\text{SimSiam}}$ and $\mathcal{A}_{\text{FedHSSL}}$ w/o and w/ ISO. This table also reports the main task performance of FedHSSL-SimSiam evaluated on 200 aligned and labeled samples. $\lambda_p$ is the noise level of ISO applied to FedHSSL pretraining. We use label recovery accuracy to measure the performance of $\mathcal{A}_{\text{SimSiam}}$ and $\mathcal{A}_{\text{FedHSSL}}$.}
 	\centering
	\begin{tabular}{l||c|c c| c c c}
	    \hline
		~ &  & \multicolumn{2}{c|}{\scriptsize{w/o ISO protection}} & \multicolumn{3}{c}{\scriptsize{w/ ISO protection}} \\
		\scriptsize{Dataset} & \scriptsize{$\mathcal{A}_{\text{SimSiam}}$} & \scriptsize{$\mathcal{A}_{\text{FedHSSL}}$} & \scriptsize{Main} & \scriptsize{$\mathcal{A}_{\text{FedHSSL}}$} & \scriptsize{Main} & $\lambda_p$  \\
	    \hline
	    \hline
	    \scriptsize{NUSWIDE} & 0.439 & 0.511 & 0.574 & 0.465 & 0.539 & 0.4 \\
	    \hline
	    Avazu & 0.545 & 0.547 & 0.616 & 0.524 & 0.617 & 0.1 \\
	    \hline
	    BHI & 0.716  & 0.726 & 0.803 & 0.682 & 0.786 & 0.1 \\
	    \hline
	    Modelnet & 0.429 & 0.441 & 0.678 & 0.426 & 0.658 & 0.1 \\
		\hline
	\end{tabular}
\label{table:privacy_pretrained_model}
\end{table}

Table \ref{table:privacy_pretrained_model} compares $\mathcal{A}_{\text{SimSiam}}$ and $\mathcal{A}_{\text{FedHSSL}}$ w/o and w/ ISO protection. Both two MC attacks leverage 80 labeled auxiliary samples to train attacking models. Table \ref{table:privacy_pretrained_model} reports that $\mathcal{A}_{\text{FedHSSL}}$ w/o ISO outperforms $\mathcal{A}_{\text{SimSiam}}$ by 0.072 on NUSWIDE and by $\leq$0.012 on the other 3 datasets, indicating that FedHSSL leaks label privacy. When ISO protection is applied with properly chosen $\lambda_p$, the performance of $\mathcal{A}_{\text{FedHSSL}}$ drops below that of $\mathcal{A}_{\text{SimSiam}}$ on 3 out of 4 datasets (except NUSWIDE), and the losses of main task performance on the 3 datasets are small ($\leq$0.02).  This manifests that the label leakage of FedHSSL can be prevented if protection mechanisms are properly applied.


\textbf{Analyzing Privacy-Utility Trade-Off.}
This experiment is conducted on the fine-tuning phase after FedHSSL pretraining. We compare FedHSSL-SimSiam with FedSplitNN and FedLocalSimSiam in terms of their privacy-utility trade-offs coming from the competition between the MC attack and the ISO protection during fine-tuning. The fine-tuning of FedHSSL-SimSiam and FedLocalSimSiam is conducted based on the two methods' pretrained models, respectively, whereas FedSplitNN involves no pretraining. For each method, Party 1 applies ISO to gradients sent back to the passive party 2 during fine-tuning/training for protection. Upon the completion of fine-tuning/training, party 2 trains a MC attacking model based on its finetuned/trained local model using $\mathcal{D}_{aux}$.

\begin{table}[!ht]
	\caption{Comparison of calibrated averaged performance (CAP) of ISO-protected FedSplitNN, FedLocalSimSiam, and FedHSSL-SimSiam against the MC attack on 4 datasets. CAP quantifies the privacy-utility trade-off curves visualized in the above 4 figures. \textit{The higher the CAP value is, the better the method is at preserving privacy without compromising the main task performances}. Numbers on the figures are values of ISO protection strength $\lambda_f$ chosen from $[1, 5, 25]$. \textit{A better trade-off curve should be more toward the bottom-right corner of each figure}. The horizontal dashed line denotes the prior knowledge of the adversary on the labels of party 1.}
\label{table:privacy_finetune_alp20}
 	\centering
  	\includegraphics[width=0.46\textwidth] {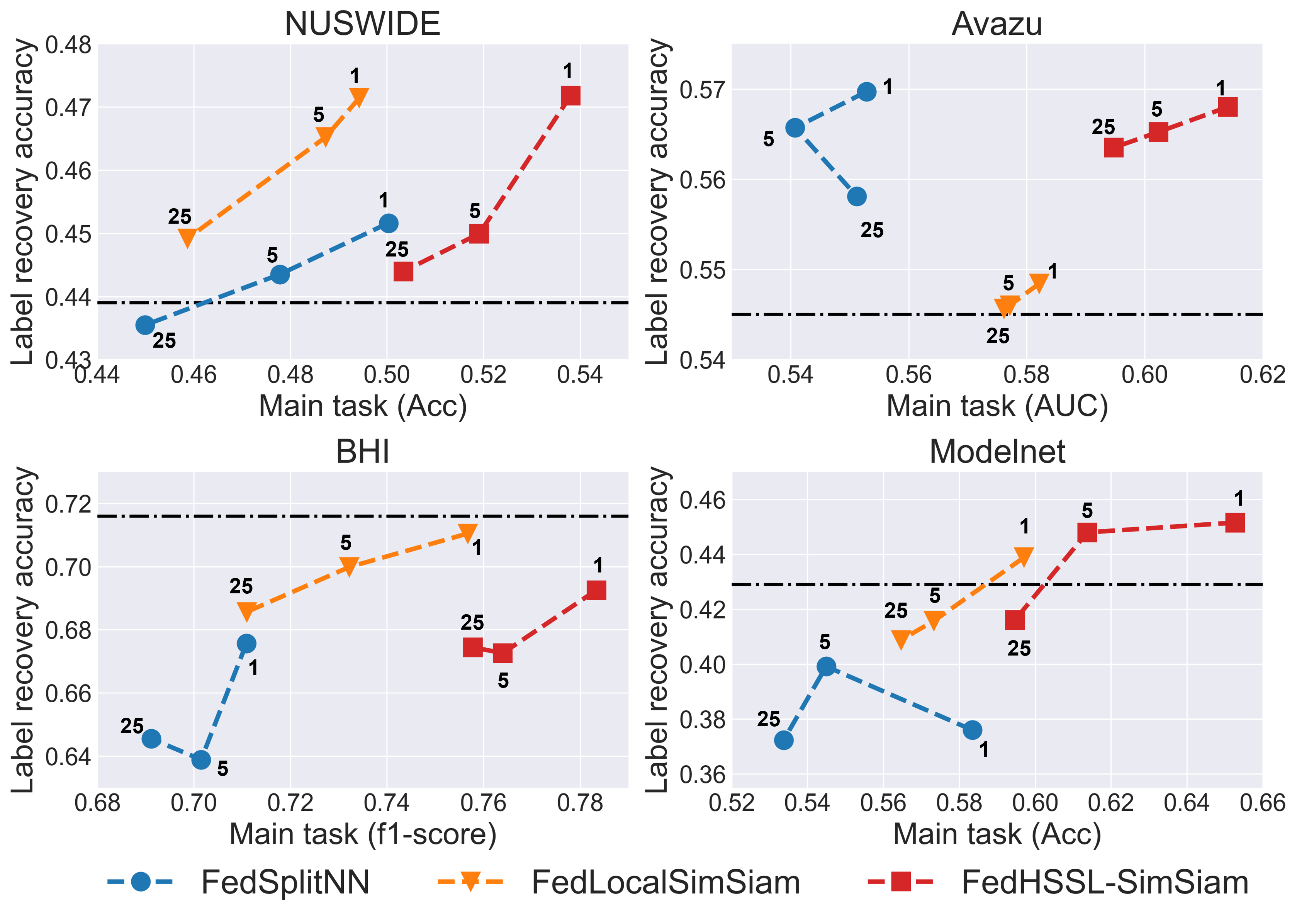}
	\begin{tabular}{l||c|c | c}
	    \hline
		\scriptsize{Dataset} & \scriptsize{FedSplitNN} & \scriptsize{FedLocalSimSiam} & \scriptsize{FedHSSL-SimSiam}  \\
	    \hline
	    \hline
	    \scriptsize{NUSWIDE} & 0.264 & 0.258 & \textbf{0.284} \\
	    \hline
     	\scriptsize{Avazu} & 0.238 & 0.262 & \textbf{0.262} \\
	    \hline
     	\scriptsize{BHI} & 0.242 & 0.221 & \textbf{0.246} \\
	    \hline
          \scriptsize{Modelnet} & 0.342 & 0.334 & \textbf{0.348} \\
		\hline
	\end{tabular}
\end{table}

From the 4 figures (in Table \ref{table:privacy_finetune_alp20}), we observe that, on each dataset, FedHSSL-SimSiam (red) achieves the best main task performance but fails to preserve the most label privacy. Thus, it is unclear whether FedHSSL-SimSiam has the best privacy-utility trade-off curve. We adopt Calibrated Averaged Performance (CAP)~\cite{Fan2020} to quantify the privacy-utility trade-off curve of a privacy-protected method so that we can compare trade-offs of different methods based on a single metric. We provide the definition of Calibrated Averaged Performance as follows.

\begin{definition}[Calibrated Averaged Performance] \label{def:cap}
For a given protection mechanism $\mathcal{M}_{\lambda}$ with a protection strength parameter $\lambda$ and an attacking mechanism $\mathcal{A}$, the Calibrated Averaged Performance (CAP) for a given privacy-utility trade-off curve is defined as follows,
\begin{equation}
    \text{CAP}(\mathcal{M}_{\lambda \in \{\lambda_1, \dots, \lambda_v \}}, \mathcal{A}) = \frac{1}{v}\sum_{\lambda=\lambda_1}^{\lambda_v} U(\bar{\mathcal{G}}_{\lambda}) * E(\bar{\mathcal{D}}_{\lambda}, \mathcal{D}),
\end{equation}
where $\bar{\mathcal{G}}_{\lambda} = \mathcal{M}_{\lambda}(\mathcal{G})$ is the VFL model protected by $\mathcal{M}_{\lambda}$, $\bar{\mathcal{D}}_{\lambda}=\mathcal{A}(\bar{\mathcal{G}}_{\lambda}, \mathcal{D})$ is the data recovered by the attacking mechanism $\mathcal{A}$ from $\bar{\mathcal{G}}_{\lambda}$ given the private data $\mathcal{D}$ as input, $U(\cdot)$ measures the main task utility (e.g., accuracy) of a given model, and $E(\cdot)$ measures the distance between recovered data $\bar{\mathcal{D}}_{\lambda}$ and original data $\mathcal{D}$.
\end{definition}

Table \ref{table:privacy_finetune_alp20} reports that, on each dataset, FedHSSL-SimSiam has the highest CAP value, and thus it achieves the best trade-off between privacy and main task performance. The reason leading to this outcome is that the amount of performance enhanced by FedHSSL-SimSiam outweighs the amount of label leakage worsened by FedHSSL-SimSiam to the extent that FedHSSL-SimSiam obtains better CAP values than baselines. With more aligned samples (i.e., 40\%) used for pretraining, FedHSSL-SimSiam generally achieves better main task performance while leaking more label privacy (see Table \ref{app_table:privacy_pretrained_model} in Appendix \ref{app_privacy}), leading to similar CAP values (see Table \ref{app_table:privacy_finetune} in Appendix \ref{app_privacy}). These experimental results manifest that the number of aligned samples is a crucial factor that impacts the privacy-utility trade-off of FedHSSL, and should be considered when applying FedHSSL.





\section{Conclusion}\label{sec:conclusion}

We propose a federated hybrid SSL framework (FedHSSL) that leverages all aligned and unaligned samples through SSL and exploits invariant features shared among parties through partial model aggregation to improve the overall performance of the VFL joint model. FedHSSL works with representative SSL methods. The experimental results show that FedHSSL outperforms baselines by a large margin. The ablation demonstrates the effectiveness of each step involved in FedHSSL. We analyze the label leakage of FedHSSL under the Model Completion (MC) attack and apply ISO to defend against MC attack. Experimental results show that FedHSSL achieves the best privacy-utility trade-off compared with baselines.

\begin{appendices}
\section{Datasets}\label{app_datasets}

\noindent\textbf{NUSWIDE} contains 634-dimensional low-level image features extracted from Flickr and 1000-dimensional corresponding text features. To simulate the VFL setting, one party holds image features, and the other holds text features. There are 81 ground truth labels, and we build datasets with our desired setting by selecting a subset of these labels. Here ten labels are for the multi-class classification task with 10 selected labels. 

\noindent\textbf{Avazu} is for predicting click-through rate. It contains 14 categorical features and 8 continuous features. We transform categorical features into embeddings with fixed dimensions (32 in this work) before feeding them the model. To simulate the VFL setting, we equally divide both kinds of features into two parts so that each party has a mixture of categorical and continuous features. To reduce the computational complexity, we randomly select 100000 samples as the training set and 20000 samples as the test set. 

\noindent\textbf{BHI (Breast Histopathology Images)} is used for binary classification task. It contains 277,524 slide-mount images of breast cancer specimens from several patients. A positive label indicates Invasive Ductal Carcinoma (IDC) positive, which is a subtype of breast cancer. The ratio between positive and negative samples is around $1:2.5$. We randomly select data of $80\%$ patients as the training set and the rest as the test set. To simulate the VFL setting, we choose two images of a patient with the same label to form a VFL sample, and each party is assigned one image. 

\noindent\textbf{Modelnet} is a multiview dataset with 40 classes. We select samples of the first 10 classes for our experiments. Each class contains several 3D objects. We generate 12 images for each object, following the procedure described in~\cite{liu2023crosssiloa}. To simulate the VFL setting, we split 12 views of each object sequentially into 4 groups so that each contains 3 nearby views, and thereby each party holds three views of an object. To expand the dataset and make the task harder, we randomly select an image from each party and build a VFL sample for each object. This procedure is the same for both the train and test sets. In the end, we have 24630 training samples and 6204 test samples.


\begin{table}[!h]
 	\centering
	\caption{Detailed information of the datasets and corresponding models.}
	\begin{tabular}{l||c|c|c|c}
	    \hline
		Dataset & Data Type &Classes & \# of Parties & Metric   \\
	    \hline
	    \hline
		NUSWIDE & Tabular & 10 & $2$ & Top-1 Acc \\
		\hline
		Avazu & Tabular & 2 & $2$ & AUC \\
		\hline
		BHI & Image & 2 & $2$ & F1-score  \\
		\hline
		Modelnet & Image & 10 & $4$ & Top-1 Acc\\
		\hline
	\end{tabular}
\label{table:data_model}
\end{table}


\section{Experimental Setup}\label{training_detail}



\subsection{Training Details}\label{app_training_detail}
For SSL training, cross-party SSL and guided local SSL are conducted alternately. Multiple epochs can be executed for both steps to reduce communication costs. In this work, we set 1 epoch for cross-party SSL and guided local SSL training. Partial model aggregation is performed directly after the guided SSL. The number of global iterations for FedHSSL prertraining is set to $10$ for NUSWIDE and $40$ for other datasets. 

All encoders include a projector consisting of 3 fully-connected layers (FC), which is only used in the pretraining phase. For FedHSSL-MoCo, the dimension of the projector is $[512, 512, 128]$. For FedHSSL-SimSiam and FedHSSL-BYOL, the dimension of the projector is $[512, 512, 512]$, and an additional 2-FC predictor with the dimension $[128, 512]$ is used. For FedHSSL-MoCo, the temperature of the InfoNCE loss is $0.5$, the size of the dictionary is 4096, and the momentum is $0.99$. For FedHSSL-BYOL, the momentum is $0.995$.

For pretraining, the batch size is $512$ for all datasets. For the finetuning, the batch size is 512 for NUSWIDE and Avazu and 128 for BHI and Modelnet. The learning rate used in the finetuning stage includes $[0.005, 0.01, 0.03]$, and the best result is selected. All experiments are repeated with 5 different seeds, and the average results are reported.


\section{More Experimental Results}\label{more_exps}

\subsection{The Impact of Cross-Party Regularization $\lambda$ on Local SSL and Model Aggregation}\label{app_reg_lambda}


We use SimSiam as the base SSL method for FedGSSL$^{*}$ and FedHSSL$^{*}$ to investigate the impact of $\gamma$. All local data and $20\%$ aligned data are used for the pretraining. 200 labeled and aligned samples are used for the finetuning.

\begin{figure}[!hb]
	\centering
	\includegraphics[width=0.45\textwidth] {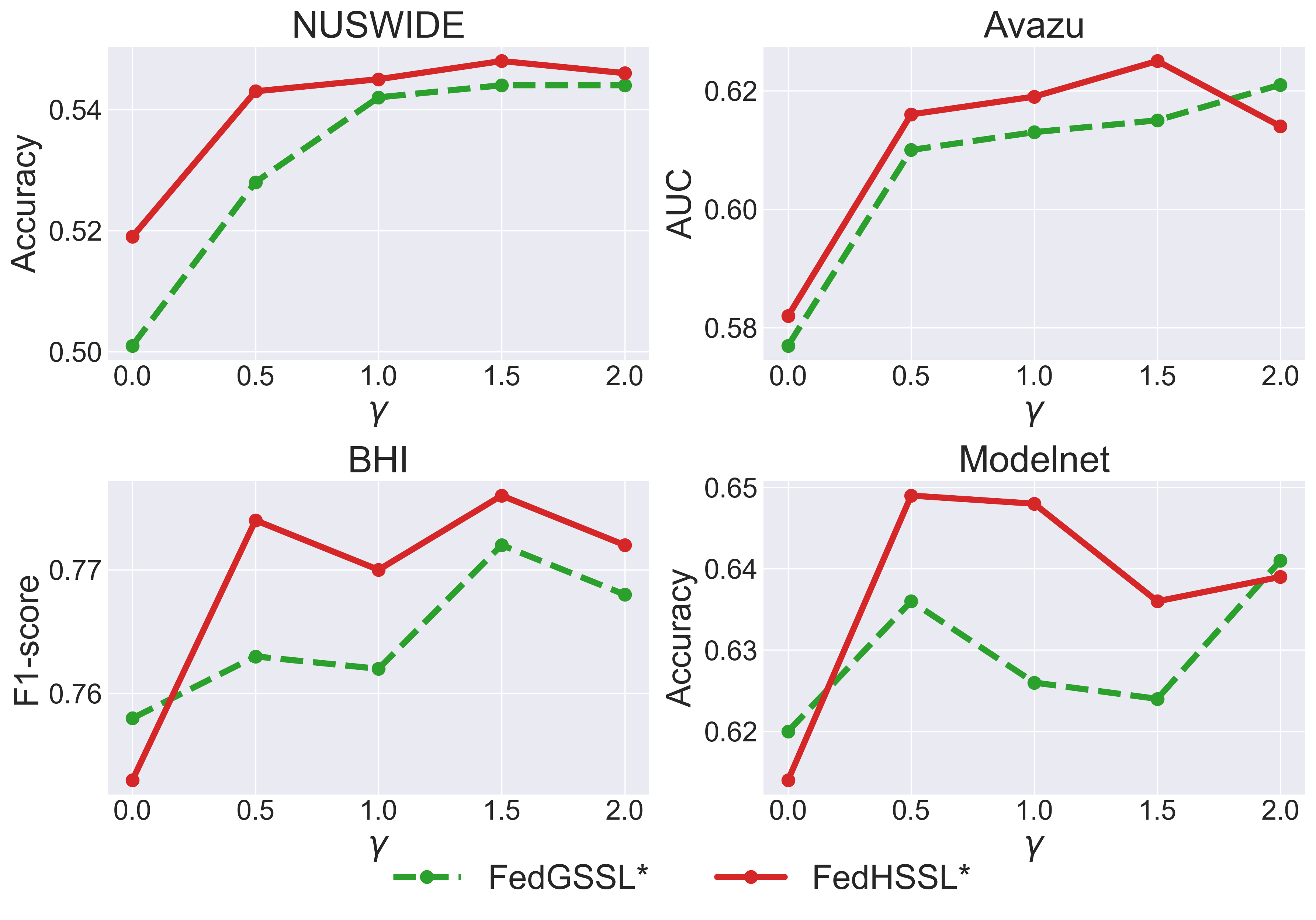}
		\vspace{-1em}
	\caption{Main task performance of FedGSSL$^{*}$ and FedHSSL$^{*}$ (use only local encoder) pretrained by various $\gamma$ values. $\gamma=0$ means no cross-party regularization is applied to local SSL.}
	\label{fig:FedGSSL_effect_app}
\end{figure}

Fig. \ref{fig:FedGSSL_effect_app} depicts the main task performance of  FedGSSL$^{*}$ and FedHSSL$^{*}$ using pretrained local encoders when $\gamma$ increases. From Fig. \ref{fig:FedGSSL_effect_app}, we observe that: i) the performance of FedGSSL$^{*}$ and FedHSSL$^{*}$ increase noticeably when $\lambda > 0$ than those of FedGSSL$^{*}$ and FedHSSL$^{*}$ when $\lambda = 0$ on four datasets, demonstrating that the cross-party regularization helps enhance the performance. ii) FedHSSL$^{*}$ constantly outperforms FedGSSL$^{*}$ on four datasets when the $\lambda$ is chosen from a proper range (i.e., 0.5 to 1.5 in this experiment), indicating that the cross-party regularization has a positive impact on the partial model aggregation when properly choosing $\lambda$. iii) the value of $\lambda$ that leads to the best performance is different for different datasets, indicating that $\lambda$ should be carefully tuned for different datasets (and models).

\begin{table*}[!ht]
 	\centering
	\caption{Performance comparison of FedCSSL-SimSiam and FedLocalSimSiam using varying percentages of training samples (\% of T.S.) for pretraining and 200 labeled samples for finetuning. }
	\begin{tabular}{l||c c c|c c c|c c c|c c c}
	    \hline
		Dataset & \multicolumn{3}{c|}{NUSWIDE} & \multicolumn{3}{c|}{Avazu} & \multicolumn{3}{c|}{BHI} & \multicolumn{3}{c}{Modelnet} \\
	     \% of T.S.: & $20\%$ & $40\%$ & $100\%$ & $20\%$ & $40\%$ & $100\%$ & $20\%$ & $40\%$ & $100\%$ & $20\%$ & $40\%$ & $100\%$ \\
	    \hline
	    \hline
		\scriptsize FedLocalSimSiam & 0.523 & 0.517 & 0.505 & 0.565 & 0.566 & 0.575 & 0.748 & 0.755 & 0.760 & 0.598 & 0.609 & 0.622  \\
		\scriptsize FedCSSL-SimSiam & \textbf{0.535} & \textbf{0.550} & \textbf{0.562} & \textbf{0.615} & \textbf{0.622} & \textbf{0.627} & \textbf{0.762} & \textbf{0.778} & \textbf{0.805} & \textbf{0.652} & \textbf{0.684} & \textbf{0.686}   \\
		\hline
        \scriptsize Enhancement &  $\uparrow$0.012 & $\uparrow$0.033 & $\uparrow$0.057 & $\uparrow$0.050 & $\uparrow$0.056 & $\uparrow$0.052 & $\uparrow$0.014 & $\uparrow$0.023 & $\uparrow$0.045 & $\uparrow$0.054 & $\uparrow$0.075 & $\uparrow$0.064 \\
        \hline
	\end{tabular}
\label{table:localss_fedss}
\end{table*}

\begin{table*}[!ht]
 	\centering
	\caption{Performance comparison of FedHSSL and baselines with different number of labeled samples. For FedHSSL, results of using $20\%$ and $40\%$ aligned samples are given. Top-1 accuracy is used as the metric for NUSWIDE and Modelnet, while AUC and F1-score are the metrics for Avazu and BHI, respectively. \% of aligned samples applies only to FedHSSL.}
	\begin{tabular}{l|l||c c|c c|c c|c c|c c}
	    \hline
		\multicolumn{2}{l||}{\# of labeled aligned samples:} & \multicolumn{2}{c|}{200} & \multicolumn{2}{c|}{400} & \multicolumn{2}{c|}{600} & \multicolumn{2}{c|}{800} & \multicolumn{2}{c}{1000}\\
		 \multicolumn{2}{l||}{\% of aligned samples:} & $20\%$ & $40\%$ & $20\%$ & $40\%$ & $20\%$ & $40\%$ & $20\%$ & $40\%$ & $20\%$ & $40\%$ \\
	    \hline
	    \hline
		\multirow{9}{*}{\shortstack{ NUSWIDE \\ (Top1-Acc)}} & LR & \multicolumn{2}{c|}{0.530} & \multicolumn{2}{c|}{0.558} & \multicolumn{2}{c|}{0.580} & \multicolumn{2}{c|}{0.589} & \multicolumn{2}{c}{0.606}    \\
		& LGB & \multicolumn{2}{c|}{0.425} & \multicolumn{2}{c|}{0.465} & \multicolumn{2}{c|}{0.526} & \multicolumn{2}{c|}{0.556} & \multicolumn{2}{c}{0.587}  \\
		& FedSplitNN & \multicolumn{2}{c|}{0.495} & \multicolumn{2}{c|}{0.535} & \multicolumn{2}{c|}{0.560} & \multicolumn{2}{c|}{0.573} & \multicolumn{2}{c}{0.591}   \\
		& FedLocalSimSiam & \multicolumn{2}{c|}{0.505} & \multicolumn{2}{c|}{0.536} & \multicolumn{2}{c|}{0.596} & \multicolumn{2}{c|}{0.603} & \multicolumn{2}{c}{0.612} \\
        & FedLocalBYOL & \multicolumn{2}{c|}{0.514} & \multicolumn{2}{c|}{0.527} & \multicolumn{2}{c|}{0.585} & \multicolumn{2}{c|}{0.599} & \multicolumn{2}{c}{0.606} \\
        & FedLocalMoCo & \multicolumn{2}{c|}{0.566} & \multicolumn{2}{c|}{0.596} & \multicolumn{2}{c|}{0.625} & \multicolumn{2}{c|}{0.634} & \multicolumn{2}{c}{0.639} \\
        \cline{2-12}
            \\[-1em]
		& \textbf{FedHSSL-SimSiam} & 0.574 & 0.607 & 0.624 & 0.641 & 0.636 & 0.651 & 0.643 & 0.662 & 0.654 & 0.670  \\
        & \textbf{FedHSSL-BYOL} & 0.551 & 0.598 & 0.592 & 0.624 & 0.617 & 0.645 & 0.633 & 0.659 & 0.640 & 0.664 \\
        & \textbf{FedHSSL-MoCo} & 0.611 & 0.615 & 0.636 & 0.642 & 0.653 & 0.658 & 0.662 & 0.668 & 0.665 & 0.670 \\
		\hline
		\multirow{9}{*}{\shortstack{Avazu \\ (AUC)}} & LR & \multicolumn{2}{c|}{0.554} & \multicolumn{2}{c|}{0.574} & \multicolumn{2}{c|}{0.596} & \multicolumn{2}{c|}{0.602} & \multicolumn{2}{c}{0.575}  \\
		& LGB & \multicolumn{2}{c|}{0.563} & \multicolumn{2}{c|}{0.568} & \multicolumn{2}{c|}{0.595} & \multicolumn{2}{c|}{0.621} & \multicolumn{2}{c}{0.620}   \\
		& FedSplitNN & \multicolumn{2}{c|}{0.588} & \multicolumn{2}{c|}{0.581} & \multicolumn{2}{c|}{0.599} & \multicolumn{2}{c|}{0.595} & \multicolumn{2}{c}{0.615}   \\
		& FedLocalSimSiam & \multicolumn{2}{c|}{0.575} & \multicolumn{2}{c|}{0.585} & \multicolumn{2}{c|}{0.591} & \multicolumn{2}{c|}{0.608} & \multicolumn{2}{c}{0.629}   \\
        & FedLocalBYOL & \multicolumn{2}{c|}{0.560} & \multicolumn{2}{c|}{0.597} & \multicolumn{2}{c|}{0.600} & \multicolumn{2}{c|}{0.601} & \multicolumn{2}{c}{0.605}   \\
        & FedLocalMoCo & \multicolumn{2}{c|}{0.573} & \multicolumn{2}{c|}{0.591} & \multicolumn{2}{c|}{0.584} & \multicolumn{2}{c|}{0.596} & \multicolumn{2}{c}{0.601}   \\
        \cline{2-12}
            \\[-1em]
		& \textbf{FedHSSL-SimSiam} & 0.616 & 0.623 & 0.625 & 0.636 & 0.631 & 0.649 & 0.644 & 0.648 & 0.657 & 0.663  \\
        & \textbf{FedHSSL-BYOL} & 0.610 & 0.615 & 0.617 & 0.634 & 0.626 & 0.631 & 0.630 & 0.630 & 0.641 & 0.648 \\
         & \textbf{FedHSSL-MoCo} & 0.614 & 0.616 & 0.623 & 0.632 & 0.635 & 0.638 & 0.637 & 0.641 & 0.646 & 0.658 \\
        \hline
		\multirow{7}{*}{\shortstack{BHI \\ (F1-Score)}} & FedSplitNN & \multicolumn{2}{c|}{0.731} & \multicolumn{2}{c|}{0.738} & \multicolumn{2}{c|}{0.754} & \multicolumn{2}{c|}{0.752} & \multicolumn{2}{c}{0.760}  \\
		& FedLocalSimSiam & \multicolumn{2}{c|}{0.760} & \multicolumn{2}{c|}{0.764} & \multicolumn{2}{c|}{0.788} & \multicolumn{2}{c|}{0.785} & \multicolumn{2}{c}{0.798}  \\
        & FedLocalBYOL & \multicolumn{2}{c|}{0.760} & \multicolumn{2}{c|}{0.769} & \multicolumn{2}{c|}{0.781} & \multicolumn{2}{c|}{0.786} & \multicolumn{2}{c}{0.796}  \\
        & FedLocalMoCo & \multicolumn{2}{c|}{0.763} & \multicolumn{2}{c|}{0.771} & \multicolumn{2}{c|}{0.784} & \multicolumn{2}{c|}{0.793} & \multicolumn{2}{c}{0.800}  \\
        \cline{2-12}
            \\[-1em]
		& \textbf{FedHSSL-SimSiam} & 0.803 & 0.805 & 0.799 & 0.816 & 0.816 & 0.822 & 0.824 & 0.823 & 0.823 & 0.830\\
        & \textbf{FedHSSL-BYOL} & 0.788 & 0.791 & 0.793 & 0.806 & 0.808 & 0.821 & 0.811 & 0.822 & 0.817 & 0.825 \\
        & \textbf{FedHSSL-MoCo} & 0.797 & 0.806 & 0.800 & 0.817 & 0.815 & 0.822 & 0.817 & 0.829 & 0.818 & 0.831 \\
		\hline
		\multirow{7}{*}{\shortstack{Modelnet \\ (Top1-Acc)}} & FedSplitNN & \multicolumn{2}{c|}{0.612} & \multicolumn{2}{c|}{0.684} & \multicolumn{2}{c|}{0.733} & \multicolumn{2}{c|}{0.765} & \multicolumn{2}{c}{0.771}   \\
		& FedLocalSimSiam & \multicolumn{2}{c|}{0.622} & \multicolumn{2}{c|}{0.698} & \multicolumn{2}{c|}{0.761} & \multicolumn{2}{c|}{0.779} & \multicolumn{2}{c}{0.797}  \\
        & FedLocalBYOL & \multicolumn{2}{c|}{0.635} & \multicolumn{2}{c|}{0.707} & \multicolumn{2}{c|}{0.760} & \multicolumn{2}{c|}{0.775} & \multicolumn{2}{c}{0.794}  \\
        & FedLocalMoCo & \multicolumn{2}{c|}{0.659} & \multicolumn{2}{c|}{0.722} & \multicolumn{2}{c|}{0.784} & \multicolumn{2}{c|}{0.798} & \multicolumn{2}{c}{0.815}  \\
        \cline{2-12}
            \\[-1em]
		& \textbf{FedHSSL-SimSiam} & 0.678 & 0.707 & 0.763 & 0.772 & 0.793 & 0.806 & 0.806 & 0.826 & 0.826 & 0.833 \\
        & \textbf{FedHSSL-BYOL} & 0.678 & 0.681 & 0.740 & 0.752 & 0.778 & 0.800 & 0.799 & 0.807 & 0.812 & 0.825 \\
        & \textbf{FedHSSL-MoCo} & 0.696 & 0.705 & 0.760 & 0.764 & 0.787 & 0.804 & 0.809 & 0.822 & 0.826 & 0.830 \\

        \hline
	\end{tabular}
\label{app_table:main_results}
\end{table*}

\subsection{Federated Cross-Party SSL vs. Local SSL in Learning Representation}
We compare the performance of FedCSSL-SimSiam and FedLocalSimSiam using varying percentages of aligned samples for SSL (i.e., 20\%, 40\%, and 100\%) and the same amount (i.e., 200) of labeled samples for finetuning. Table \ref{table:localss_fedss} reports that FedCSSL-SimSiam outperforms FedLocalSimSiam on all sample percentages across all datasets. With more samples used for pretraining (from $20\%$ to $100\%$), the performance improvement becomes larger, especially on NUSWIDE (by $0.045$) and BHI (by $0.031$). This demonstrates that FedCSSL-SimSiam is more effective in pretraining representation than FedLocalSimSiam, indicating that the features (cross-party views) of aligned samples form better positive pairs for the SSL than the local augmentation. These experiments prove the merit of VFL in building better machine learning models.

\subsection{The Impact of the Amount of Aligned Samples on FedHSSL}\label{app_more_data}
We compare the performance of FedHSSL using various amount of aligned samples, $20\%$ and $40\%$ respectively. The results in Table \ref{app_table:main_results} show that the performance of FedHSSL improves constantly with more aligned samples. This suggests that more aligned samples help FedHSSL generate better representations for downstream tasks.




\begin{table*}[!h]
	\caption{Comparison of MC attack (privacy leakage) vs. main task (utility) trade-offs for ISO-protected  FedLocalSimSiam and FedHSSL-SimSiam on 4 datasets with $20\%$ and $40\%$ aligned samples, respectively. $\lambda_f$ indicates the protection strength used in the finetuning phase and $\lambda_p$ the protection strength in the pretraining phase.}
 	\centering
	\footnotesize
	\begin{tabular}{l||c|c c|c c c| c c c}
	    \hline
		Method &  & \multicolumn{2}{c|}{FedLocalSimSiam} & \multicolumn{3}{c|}{FedHSSL-SimSiam ($20\%$)} & \multicolumn{3}{c}{FedHSSL-SimSiam ($40\%$)} \\
		\scriptsize{Dataset} & $\lambda_f$ & \scriptsize{$\mathcal{A}_{\text{SimSiam}}$} & \scriptsize{Main} & \scriptsize{$\mathcal{A}_{\text{FedHSSL}}$} & \scriptsize{Main} & $\lambda_p$ & \scriptsize{$\mathcal{A}_{\text{FedHSSL}}$} & \scriptsize{Main} & $\lambda_p$  \\
	    \hline
	    \hline
	    \multirow{3}{*}{\scriptsize{NUSWIDE}} & 1.0 & 0.471 & 0.494 & 0.471 & 0.538 & 0.4 & 0.474 & 0.533 & 5.0 \\ 
	    & 5.0 & 0.465 & 0.487 &  0.449 & 0.519 & 0.4 & 0.471 & 0.528 & 5.0 \\
	    & 25.0 & 0.449 & 0.458 & 0.443 & 0.503 & 0.4 & 0.458 & 0.503 &  5.0 \\
	    \hline
	    \multirow{3}{*}{Avazu} & 1.0 & 0.548 & 0.582 & 0.568 & 0.614 & 0.1 & 0.571 & 0.616 & 0.1 \\
	    & 5.0 & 0.546 & 0.577 & 0.565 & 0.602 & 0.1 & 0.566 & 0.610 & 0.1 \\
	    & 25.0 & 0.545 & 0.576 & 0.563 & 0.594 & 0.1 & 0.561 & 0.603 & 0.1 \\
	    \hline
	    \multirow{3}{*}{BHI} & 1.0 & 0.710 & 0.756& 0.692 & 0.783 & 0.1 &  0.686 & 0.788 & 0.1 \\
	    & 5.0 & 0.699 & 0.732 & 0.672 & 0.764 & 0.1&  0.687 & 0.773 & 0.1 \\
	    & 25.0 & 0.685 & 0.710 & 0.674 & 0.758 & 0.1&  0.682 & 0.764 & 0.1 \\
	    \hline
	    \multirow{3}{*}{Modelnet} & 1.0 & 0.438 & 0.597 & 0.451 & 0.652 & 0.1 &  0.466 & 0.658 & 0.1 \\
	    & 5.0 & 0.415 & 0.573 & 0.447 & 0.613 & 0.1&  0.448 & 0.631 & 0.1 \\
	    & 25.0 & 0.408 & 0.564 & 0.415 & 0.594 & 0.1&  0.419 & 0.598 & 0.1 \\
		\hline
	\end{tabular}
\label{app_table:privacy_pretrained_model}
\end{table*}


\begin{table*}[ht]
	\caption{Comparison of calibrated averaged performance (CAP) of ISO-protected FedSplitNN, FedLocalSimSiam and FedHSSL-SimSiam against the MC attack on 4 datasets. CAP quantifies the privacy-utility trade-off curves visualized in above 4 figures. \textit{The higher the CAP value is, the better the method is at preserving privacy without compromising the main task performances}. Numbers on the figures are values of ISO protection strength $\lambda_f$ chosen from $[1, 5, 25]$. A better trade-off curve should be more towards the bottom-right corner of each figure. }
 	\centering
  	\includegraphics[width=0.45\textwidth] {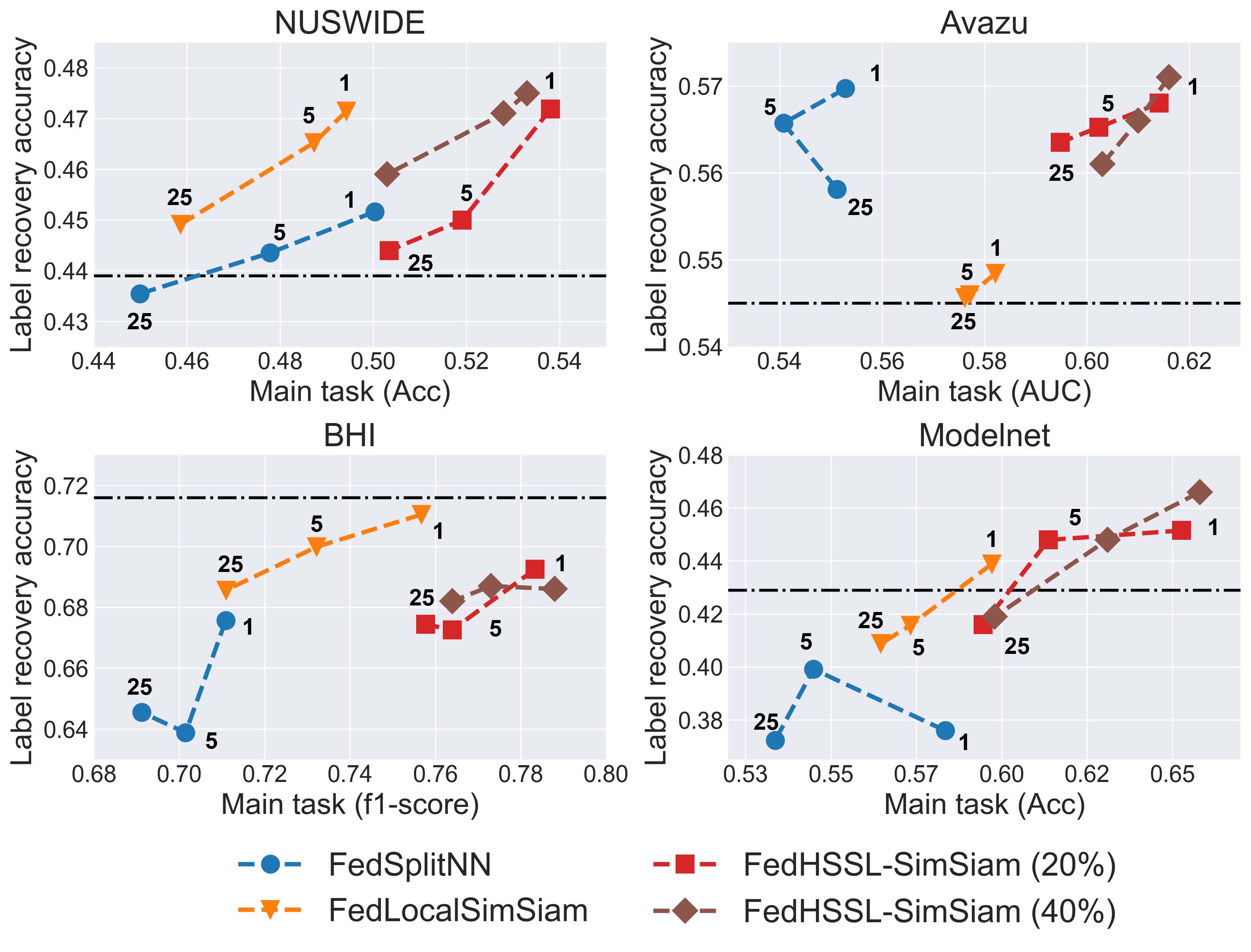}
	\scriptsize
	\begin{tabular}{l||c|c | c | c}
	    \hline
		\scriptsize{Dataset} & \scriptsize{FedSplitNN} & \scriptsize{FedLocalSimSiam} & \scriptsize{FedHSSL-SimSiam ($20\%$)} &\scriptsize{FedHSSL-SimSiam ($40\%$)} \\
	    \hline
	    \hline
	    \scriptsize{NUSWIDE} & 0.264 & 0.258 & \textbf{0.284} & \textbf{0.277}\\
	    \hline
     	\scriptsize{Avazu} & 0.238 & 0.262 & \textbf{0.262}  & \textbf{0.264} \\
	    \hline
     	\scriptsize{BHI} & 0.242 & 0.221 & \textbf{0.246}   & \textbf{0.244}\\
	    \hline
          \scriptsize{Modelnet} & 0.342 & 0.334 & \textbf{0.348}  & \textbf{0.349} \\
		\hline
	\end{tabular}
\label{app_table:privacy_finetune}
\end{table*}

\subsection{Privacy Analysis Of FedHSSL with Different Aligned Samples}\label{app_privacy}
We investigate the privacy-utility trade-off of FedHSSL in terms of various amount of aligned samples. We use SimSiam as the base SSL method for FedHSSL. As shown in Table \ref{app_table:privacy_pretrained_model}, with more aligned samples (i.e., from $20\%$ to $40\%$) are used for pretraining, the main task performance of FedHSSL-SimSiam is generally improved while the label recovery accuracy is also increasing when the same level of protection strength is applied. This trends is also illustrated in figures of Table \ref{app_table:privacy_finetune}, which reports that, while FedHSSL-SimSiam gives different privacy-utility trade-off curves when leveraging different amount of aligned samples, the two curves have similar CAP values. This result manifests that the number of aligned samples is an important factor that impacts the privacy-utility trade-off of FedHSSL, and should be considered when applying FedHSSL.


\end{appendices}

\bibliographystyle{IEEEtran}
\bibliography{IEEEabrv,main}

\newpage

 



\vfill

\end{document}